\documentclass[sigconf]{acmart}

\AtBeginDocument{%
  }

\copyrightyear{2026}
\acmYear{2026}
\setcopyright{cc}
\setcctype{by}
\acmConference[MM '26]{Proceedings of the 34th ACM International Conference on Multimedia}{November 10--14, 2026}{Rio de Janeiro, Brazil}
\acmBooktitle{Proceedings of the 34th ACM International Conference on Multimedia (MM '26), November 10--14, 2026, Rio de Janeiro, Brazil}
\acmDOI{10.1145/3767308.3835539}
\acmISBN{979-8-4007-2213-4/2026/11}

\usepackage{balance}
\usepackage{subcaption}
\usepackage{multirow, booktabs, graphicx}
\usepackage{makecell}
\usepackage{graphicx}
\usepackage{amsfonts}
\usepackage{amsmath}
\usepackage{multirow}
\usepackage{xcolor}
\usepackage{booktabs}
\usepackage{algorithm}
\usepackage{algpseudocode}
\usepackage{setspace}
\usepackage{cleveref}
\usepackage{enumitem}

\usepackage{pifont}

\begin{document}

\title{DriftAD: Visually-Guided Text Drift for Few-Shot Industrial Anomaly Detection}

\author{Wenyang Liu}
\email{wenyang001@e.ntu.edu.sg}
\affiliation{%
  \institution{Nanyang Technological University}
  \country{Singapore}
}

\author{Tianyi Liu}
\email{liut0038@e.ntu.edu.sg}
\affiliation{%
  \institution{Nanyang Technological University}
  \country{Singapore}
}

\author{Dongshuo Zhang}
\email{dongshuo001@e.ntu.edu.sg}
\affiliation{%
  \institution{Nanyang Technological University}
  \country{Singapore}
}

\author{Kejun Wu}
\email{kjwu@hust.edu.cn}
\affiliation{%
  \institution{Huazhong University of Science and Technology, Wuhan, China}
  \country{}
}

\author{Adams Wai-Kin Kong}
\email{adamskong@ntu.edu.sg}
\affiliation{%
  \institution{Nanyang Technological University}
  \country{Singapore}
}


\renewcommand{\shortauthors}{Wenyang Liu, Tianyi Liu, Dongshuo Zhang, Kejun Wu, and Adams Wai-Kin Kong}

\begin{abstract}
Few-shot anomaly detection (FSAD) has recently benefited from vision-language models such as CLIP, which enable anomaly detection by aligning visual features with text descriptions of normal and abnormal states.
However, existing methods typically rely on static text prompts that are applied uniformly across the entire feature hierarchy and spatial dimensions. This rigid global-to-local matching fails to capture the highly localized and scale-dependent physical variations of industrial defects.
To address this, we propose DriftAD, a FSAD framework built on three key modules.
First, an Anomaly Signal Amplification (ASA) module enhances subtle defect signals
through spatial and frequency branches before text-visual matching.
Second, Visually-Guided Text Drift (VGTD) dynamically transforms frozen CLIP text embeddings, steering them into layer-wise, spatially-adaptive anomaly descriptors conditioned on local visual context at each encoder depth.
Third, Drift-Guided Spatial Gating (DGSG) uses the drifted abnormal descriptor as a
spatial probe to selectively enhance anomaly-relevant visual features.
Additionally, a drift separation loss prevents representational collapse of the drifted descriptors,
and a gate supervision loss enforces spatially discriminative gating in DGSG.
Extensive experiments on MVTec-AD and VisA demonstrate state-of-the-art performance
across all 1-, 2-, and 4-shot settings on both image-level and pixel-level metrics. Code is available at
\url{https://github.com/wenyang001/DriftAD}.
\end{abstract}




\maketitle

\begin{figure}[t]
    \centering
\includegraphics[width=0.95\columnwidth]{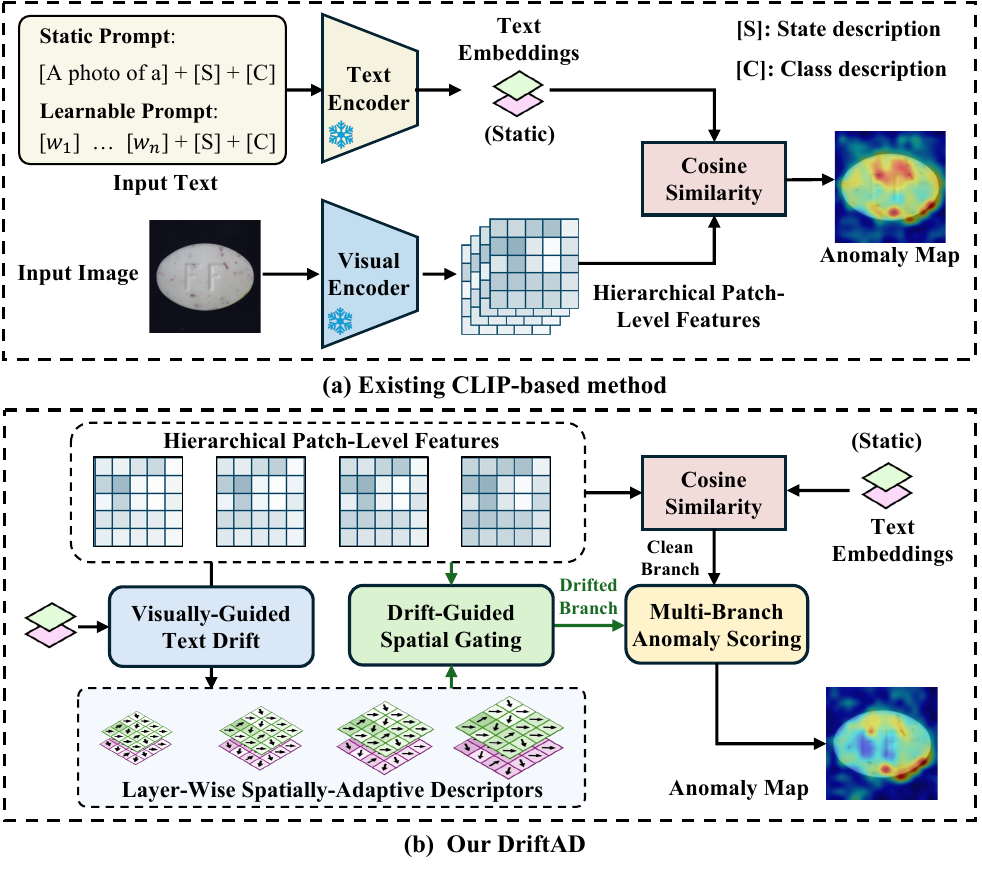}
    \vspace{-0.2in}
\caption{(a) Existing approaches mainly rely on globally shared, static text representations (even when utilizing prompt-tuning), inherently restricting their sensitivity to localized defects. (b) DriftAD treats these static representations strictly as initial anchors. By incorporating local visual context, it dynamically drifts these anchors into layer-wise, spatially-adaptive anomaly descriptors, achieving significantly sharper and more precise anomaly maps.}
\label{fig:motivation}
    \vspace{-0.2in}
\end{figure}

\vspace{-0.05in}
\section{Introduction}

Industrial anomaly detection (AD) aims to localize rare defects against abundant normal patterns. This task has long been dominated by unsupervised anomaly detection (UAD) methods such as CFLOW-AD~\cite{gudovskiy2022cflow}, PaDiM~\cite{defard2021padim}, and PatchCore~\cite{roth2022towards}. While these approaches achieve strong performance by modeling normality through density estimation or memory banks, they are constrained by a ``one-model-per-category'' paradigm, limiting their scalability across diverse product categories.
This limitation has motivated a shift toward open-set generalization by exploiting the rich semantic priors of vision-language models (VLMs) such as CLIP~\cite{radford2021learning}. WinCLIP~\cite{jeong2023winclip} pioneers this direction by computing multi-scale patch-text similarity using hand-crafted state-class templates, where \textit{state} denotes the object condition (e.g., ``normal'' or ``damaged'') and \textit{class} denotes the object category (e.g., ``bottle'' or ``cable''), yielding descriptors such as ``a photo of a \{state\} \{class\}''. Following this direction, AnomalyGPT~\cite{gu2024anomalygpt} incorporates large language models for interactive anomaly reasoning, while PromptAD~\cite{li2024promptad} replaces hand-crafted prompts with learnable continuous vectors optimized on few-shot support images, producing category-adaptive descriptors of the form ``$[\mathbf{w}_1,\ldots,\mathbf{w}_E]$ \{state\} \{class\}''. In parallel, KAG-prompt~\cite{tao2025kernel} strengthens visual representations through kernel-aware graph neural networks that model cross-layer feature interactions, while FocusPatch-AD~\cite{ding2025focuspatch} associates anomaly-related keywords with spatially relevant regions to suppress background interference. Together, these advances improve adaptation and generalization to unseen categories with limited supervision, reducing reliance on category-specific models and enabling more flexible and scalable anomaly detection.


Despite this progress, existing VLM-based methods share a fundamental limitation: the anomaly text embedding remains a static, category-level global descriptor. As illustrated in Figure~\ref{fig:motivation}(a), this fixed representation is shared uniformly across all spatial locations and encoder depths, regardless of the input image content. Methods that refine the text side~\cite{jeong2023winclip,li2024promptad} optimize the descriptor at the category level but cannot adapt it to individual test images, while methods that strengthen the visual side~\cite{tao2025kernel,ding2025focuspatch} leave the text descriptor entirely frozen. In both cases, the text embedding lacks awareness of the precise location and morphology of a defect in the current image, fundamentally limiting its ability to serve as a precise spatial reference. This design leads to diffuse anomaly maps that struggle to distinguish fine-grained defect regions from surrounding normal patterns. This static design mismatches the intrinsic nature of industrial defects: while normal patterns are relatively stable, anomalies are inherently diverse, context-dependent, and manifest as distinct localized deviations.

We present DriftAD to address this limitation by introducing Visually-Guided Text Drift (VGTD), which transforms frozen CLIP text embeddings into layer-wise, spatially-adaptive anomaly descriptors conditioned on the local visual context of each input image.
Rather than a fixed category-level token, the normal and abnormal descriptors dynamically drift from their corresponding frozen text embeddings based on the visual features at each spatial
location and encoder depth, as illustrated in Figure~\ref{fig:motivation}(b).
Concretely, DriftAD first employs an Anomaly Signal Amplification (ASA) module to
amplify anomaly-relevant signals suppressed by the dominant normal pattern through
spatial and frequency branches.
Guided by these amplified features, VGTD generates spatially-varying drift fields at
each encoder depth that displace the frozen text embeddings into spatially-adaptive
anomaly descriptors.
The drifted abnormal descriptor then serves as a spatial probe within Drift-Guided
Spatial Gating (DGSG), selectively enhancing visual features with high anomaly affinity
prior to text-visual matching.
To ensure robust optimization, DriftAD is trained with a drift separation loss that
prevents representational collapse of the drifted descriptors and a gate supervision
loss that enforces spatially discriminative gating in DGSG.
Extensive experiments on MVTec-AD and VisA demonstrate state-of-the-art performance across all few-shot settings, outperforming existing prompt-based and visual-enhancement methods on almost all metrics. In summary, our main contributions are:

\begin{itemize}
    \item We present DriftAD, a few-shot anomaly detection method that
    rethinks the anomaly text representation: rather than a fixed global token, DriftAD
    introduces Visually-Guided Text Drift (VGTD) to shift frozen CLIP embeddings
    along visually-guided directions, generating layer-wise, spatially-adaptive
    anomaly descriptors conditioned on each input image.

    \item We propose VGTD, which transforms frozen CLIP embeddings into layer-wise, spatially-adaptive anomaly descriptors via dynamic drift fields conditioned on local visual context, and
    Drift-Guided Spatial Gating (DGSG), which uses the drifted descriptor as a spatial
    probe to selectively enhance anomaly-relevant features.
    An Anomaly Signal Amplification (ASA) module additionally exposes subtle defect
    signals through spatial and frequency branches before matching.

    \item To further regularize the proposed modules, we introduce a gate supervision loss
that enforces spatially discriminative attention in DGSG, and a drift separation loss
that prevents the drifted normal and abnormal descriptors in VGTD from collapsing
toward similar representations.
    Extensive experiments on MVTec-AD and
    VisA demonstrate state-of-the-art performance on both image-level and pixel-level metrics.
\end{itemize}

\section{Related Work}

\subsection{Anomaly Detection}
Industrial anomaly detection aims to identify and localize visual deviations, such as surface defects and structural flaws, in manufactured products, making it essential for automated quality control. Recent advances in visual representation learning have improved the modeling of complex appearances, local structures, and semantic information across various vision tasks~\cite{liu2023bitstream,liu2026promptsr,radford2021learning,zhou2023anomalyclip}. Nevertheless, anomaly detection remains challenging because real defects are rare and unpredictable. Conventional methods mainly learn normal patterns from defect-free data and identify anomalies as deviations from learned normality. Embedding-based methods, such as PaDiM~\cite{defard2021padim}, PatchCore~\cite{roth2022towards}, CFLOW-AD~\cite{gudovskiy2022cflow}, and SimpleNet~\cite{liu2023simplenet} detect anomalies by measuring deviations from nominal feature distributions. Distillation-based methods, including RD4AD~\cite{deng2022anomaly}, and MemKD~\cite{gu2023remembering}, exploit teacher-student feature discrepancies to reveal abnormal regions. Reconstruction-based approaches identify anomalies from discrepancies between inputs and reconstructed normal counterparts, with recent methods adopting diffusion models such as AnoDDPM~\cite{wyatt2022anoddpm} and DiAD~\cite{he2024diffusion}.

Despite achieving strong performance, these traditional unsupervised paradigms share a critical limitation: they require training a separate, dedicated model for each specific object category using extensive normal samples. 
This closed-set, per-category training requirement incurs significant computational and memory costs. 
Consequently, their scalability and efficiency are severely limited in dynamic, real-world industrial applications, motivating the recent shift toward open-set and few-shot detection frameworks.

\subsection{Zero- and Few-Shot Anomaly Detection}

To overcome the closed-set limitations of traditional paradigms, recent research has rapidly shifted toward open-set anomaly detection under limited data regimes. Vision-language models (VLMs) such as CLIP~\cite{radford2021learning} have established themselves as the mainstream backbone for extracting rich semantic priors. WinCLIP~\cite{jeong2023winclip} pioneers this direction in the zero-shot setting by matching multi-scale visual patches with manually designed text prompts. Subsequent works, including AnomalyCLIP~\cite{zhou2023anomalyclip} and AdaCLIP~\cite{cao2024adaclip}, extend this framework by introducing learnable continuous prompts and dedicated projectors to better adapt CLIP to anomaly detection. However, zero-shot approaches inherently struggle to capture fine-grained, domain-specific features without any visual reference, motivating few-shot anomaly detection (FSAD), which leverages a small number of normal support images to bridge this gap.

While early FSAD studies such as RegAD~\cite{huang2022registration} and FastRecon~\cite{fang2023fastrecon} focused on feature transformation and distribution regularization, recent methods leverage limited visual references to improve cross-category generalization.
NAGL~\cite{wang2026normal} introduces normal-abnormal guided learning for generalist anomaly detection, while IIPAD~\cite{lv2025one} learns instance-induced continuous prompts and AnomalyGPT~\cite{gu2024anomalygpt} employs a visual-textual decoder for pixel-level localization. While these advances significantly improve open-set generalization, precise alignment between visual and textual modalities remains a fundamental bottleneck.


\begin{figure*}[t]
    \centering
\includegraphics[width=0.9\textwidth]{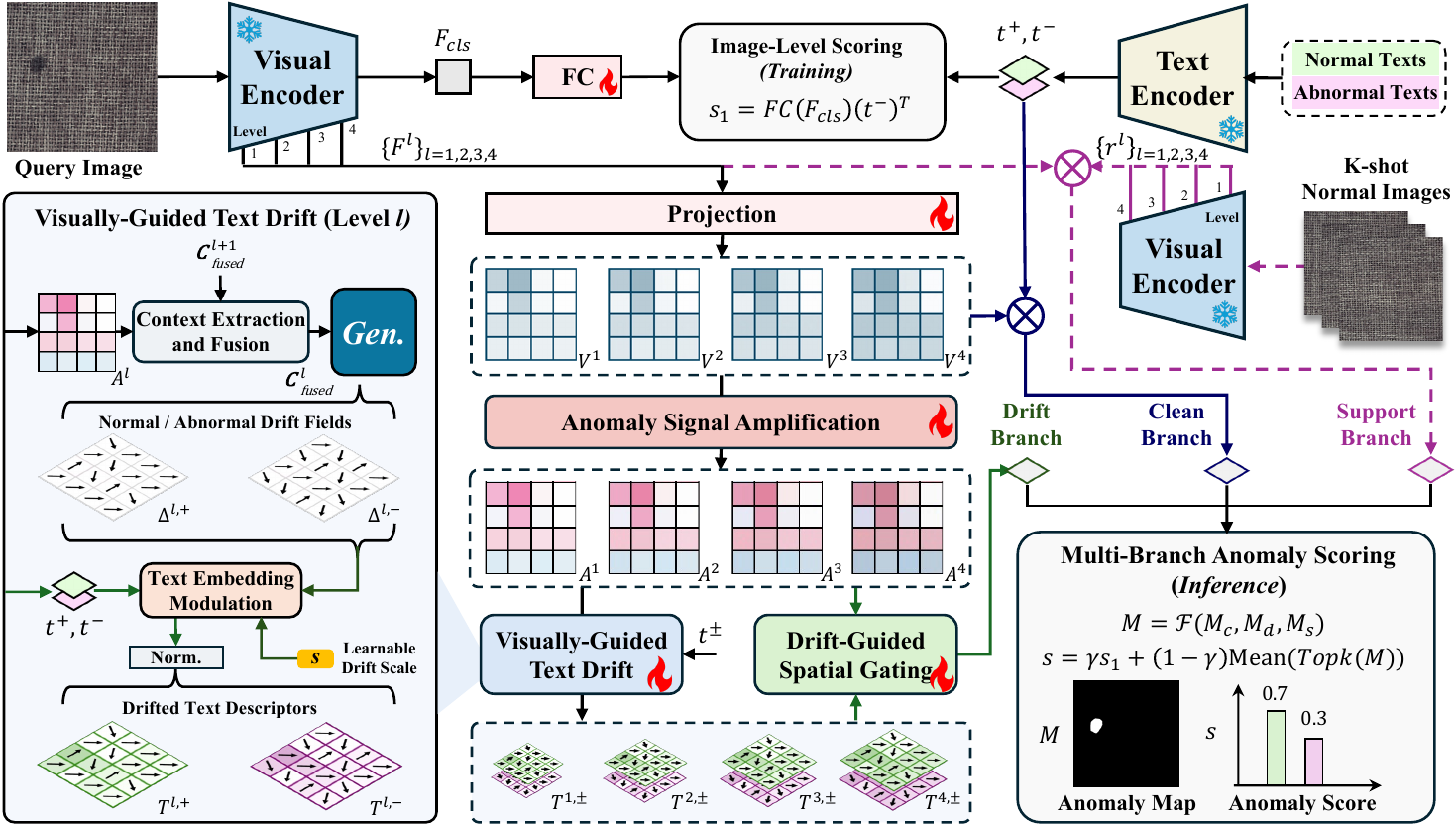}
    \vspace{-0.1in}
\caption{Overview of our method. Given a query image, the frozen image encoder extracts multi-level patch features ${F^\ell}$ and a CLS token $F_{\text{cls}}$, which are projected into scale-aware features ${V^\ell}$ and enhanced by Anomaly Signal Amplification to obtain ${A^\ell}$. Visually-Guided Text Drift conditions spatially-varying drift fields on ${A^\ell}$ to transform frozen text embeddings into layer-wise, spatially-adaptive anomaly descriptors ${T^{\ell,\pm}}$. Drift-Guided Spatial Gating then uses $T^{\ell,-}$ as a spatial probe to enhance anomaly-relevant regions. Multi-Branch Anomaly Scoring fuses three complementary branches to produce the final anomaly map $M$ and score $s$. Dashed lines denote the Support Branch, activated only at inference.}
\label{fig:overview}
\vspace{-0.15in}
\end{figure*}

\subsection{Text-Visual Anomaly Alignment}
To address this alignment bottleneck, existing methods mainly pursue two complementary directions. On the text side, inspired by CoOp~\cite{zhou2022learning} and PromptSRC~\cite{khattak2023self}, methods such as PromptAD~\cite{li2024promptad} and AnomalyCLIP~\cite{zhou2023anomalyclip} replace hand-crafted templates with learnable continuous prompt tokens. However, these prompts are global category-level parameters fixed at test time, and thus cannot adapt to the local visual content of each query image.

On the visual side, APRIL-GAN~\cite{chen2023april} projects patch features into the text space via learnable linear layers, KAG-prompt~\cite{tao2025kernel} aggregates cross-layer visual features through kernel-aware graph-based message passing, and FocusPatch-AD~\cite{ding2025focuspatch} links anomaly state keywords to spatially relevant local regions to suppress background interference. While these methods strengthen the visual representation before matching, the text descriptor itself remains a fixed global token with no knowledge of the input image, limiting the precision of patch-level text-visual alignment. DriftAD addresses this asymmetry by dynamically conditioning the text descriptor on the local visual context at each spatial position and encoder depth, producing layer-wise, spatially adaptive anomaly descriptors that adapt to the specific defect pattern of each input image.

\section{Methodology}

An overview of DriftAD is shown in Figure~\ref{fig:overview}. Given a query image $x \in \mathbb{R}^{3 \times H \times W}$, the frozen image encoder extracts patch tokens $F^\ell \in \mathbb{R}^{N \times D_0}$ and a CLS token $F_{\text{cls}} \in \mathbb{R}^{D_0}$ from four hierarchical levels $\ell \in \{1,2,3,4\}$, where $N=256$ and $D_0=1280$. Following~\cite{jeong2023winclip, tao2025kernel}, the patch tokens are projected and aggregated with multi-kernel convolutions to obtain scale-aware features $V^\ell \in \mathbb{R}^{D \times H_s \times W_s}$, where $D=1024$ and $H_s=W_s=16$. In parallel, the class name $c$ is encoded by the frozen text encoder with prompt ensembling to obtain normal and abnormal embeddings $\mathbf{t}^{+}, \mathbf{t}^{-} \in \mathbb{R}^{D}$.

The pipeline consists of four components.
\textbf{(i) Anomaly Signal Amplification (ASA):} $V^\ell$ is enhanced through spatial and frequency branches to obtain $A^\ell$, exposing anomaly-relevant signals suppressed by dominant normal patterns.
\textbf{(ii) Visually-Guided Text Drift (VGTD):} ${A^\ell}$ and ${\mathbf{t}^{+}, \mathbf{t}^{-}}$ are jointly fed into VGTD to generate layer-wise, spatially-adaptive anomaly descriptors $T^{\ell,\pm}$ through context extraction, fusion, and drift-based modulation.
\textbf{(iii) Drift-Guided Spatial Gating (DGSG):} $T^{\ell,-}$ serves as a spatial probe to generate a multi-head similarity gate over $A^\ell$, enhancing anomaly-relevant regions and producing refined features $\tilde{A}^\ell$ via skip-connected fusion.
\textbf{(iv) Multi-Branch Anomaly Scoring (MBAS):} $V^\ell$, $\tilde{A}^\ell$, and $F^\ell$ are matched with their respective text references across three complementary branches to produce the final anomaly map $M$ and image-level score $s$.

\subsection{Anomaly Signal Amplification}

Given the projected features $V^\ell$, subtle defect signals are easily suppressed by the
dominant normal pattern before text-visual matching.
We introduce ASA to amplify anomaly-relevant signals in $V^\ell$: the spatial branch
highlights deviating regions by contrasting each feature with its local background, while
the frequency branch enhances abnormal spectral components by adaptively reweighting the
frequency spectrum of the feature map.

The spatial branch computes multi-scale residuals between $V^\ell$ and its smoothed
backgrounds via $3{\times}3$ and $9{\times}9$ average pooling to obtain $R_s$, which is
then passed through a lightweight attention network to produce a spatial amplification
weight $\omega_s = \sigma(\mathrm{Conv}(\mathrm{Conv}(R_s)))$ that modulates $V^\ell$:
\begin{gather}
    R_s = |V^\ell - \mathrm{AvgPool}_{3\times3}(V^\ell)| +
    |V^\ell - \mathrm{AvgPool}_{9\times9}(V^\ell)|, \label{eq:residual} \\
    \mathbf{A}^\ell_{\text{spa}} = V^\ell + V^\ell \odot \omega_s \cdot \alpha,
    \label{eq:spatial}
\end{gather}
where $\alpha$ is a learnable scale parameter.

The frequency branch decomposes $V^\ell$ into magnitude and phase spectra via 2D FFT,
$\mathbf{E} = |\mathrm{FFT}(V^\ell)|$ and $\boldsymbol{\Phi} = \angle\mathrm{FFT}(V^\ell)$,
and reweights the low- and high-frequency components with learnable weights $\omega_l$
and $\omega_h$:
\begin{gather}
    \mathbf{E}' = \omega_l \cdot \mathbf{E}_{\text{low}} + \omega_h \cdot \mathbf{E}_{\text{high}},
    \label{eq:freq} \\
    \mathbf{A}^\ell_{\text{freq}} = \mathrm{Conv}\!\left([V^\ell,\; |V^\ell -
    \mathrm{IFFT}(\mathbf{E}' \cdot e^{j\boldsymbol{\Phi}})|]\right),
    \label{eq:freq_out}
\end{gather}
where $\mathrm{IFFT}(\mathbf{E}' \cdot e^{j\boldsymbol{\Phi}})$ denotes the
frequency-reconstructed feature, and the residual highlights regions whose frequency
content deviates from the normal pattern.

The two branches are adaptively combined with a learnable weight $\beta \in [0, 1]$:
\begin{equation}
    A^\ell = \beta \cdot \mathbf{A}^\ell_{\text{spa}} + (1 - \beta) \cdot \mathbf{A}^\ell_{\text{freq}},
    \label{eq:amp_fusion}
\end{equation}
allowing the model to balance structural and textural anomaly signals across different
defect categories.
\subsection{Visually-Guided Text Drift}

Existing FSAD methods treat the anomaly class as a single global text embedding
$\mathbf{t}^{-} \in \mathbb{R}^{D}$, applied uniformly across all spatial positions and
encoder depths, failing to capture the spatially localized and layer-dependent nature of
industrial defects.
We propose VGTD, which transforms the static embeddings $\mathbf{t}^{+}, \mathbf{t}^{-}$
into layer-wise, spatially-adaptive anomaly descriptors $T^{\ell,\pm} \in \mathbb{R}^{D \times H_s
\times W_s}$ by conditioning the drift on the amplified visual features $A^\ell$ at each
encoder depth.

\vspace{-0.5ex}
\paragraph{Context Extraction and Fusion.}
At each layer $\ell$, a dual visual context is extracted from $A^\ell$ via adaptive average
and max pooling to a layer-specific grid $g_\ell \in \{4, 8, 8, 16\}$, and projected to
obtain the raw context $\mathbf{C}^\ell_{\text{raw}} \in \mathbb{R}^{D \times g_\ell \times g_\ell}$:
\begin{equation}
    \mathbf{C}^\ell_{\text{raw}} = \mathrm{Conv}\left(
    [\mathrm{AvgPool}_{g_\ell}(A^\ell),\; \mathrm{MaxPool}_{g_\ell}(A^\ell)]\right).
    \label{eq:raw_ctx}
\end{equation}
Since shallow layers lack the semantic richness needed to produce meaningful drift, we
propagate context from deep to shallow layers via top-down fusion, starting from $\ell = 4$
and iteratively merging into each shallower layer:
\begin{equation}
    \mathbf{C}^\ell_{\text{fused}} = \mathrm{Conv}\left([\mathbf{C}^\ell_{\text{raw}},\;
    \mathbf{C}^{\ell+1}_{\text{fused}} \downarrow_{g_\ell}]\right), \quad \ell = 3, 2, 1,
    \label{eq:topdown}
\end{equation}
where $\downarrow_{g_\ell}$ denotes bilinear interpolation to the target grid size $g_\ell$,
ensuring that even the coarsest $4{\times}4$ drift field is informed by the global semantic
context of deeper layers.

\vspace{-0.5ex}
\paragraph{Drift Embedding Modulation.}
Given $\mathbf{C}^\ell_{\text{fused}}$, each layer maintains a dedicated generator
$\mathrm{Gen}$ with independent parameters to produce normal and abnormal drift fields:
\begin{equation}
    \triangle^{\ell,+},\, \triangle^{\ell,-} = \mathrm{Gen}(\mathbf{C}^\ell_{\text{fused}}),
    \label{eq:delta}
\end{equation}
where $\triangle^{\ell,\pm} \in \mathbb{R}^{D \times g_\ell \times g_\ell}$ are
$\ell_2$-normalized and scaled by a learnable drift scale $\rho$ to constrain the
displacement within the CLIP feature space.
The drifted text descriptors are obtained by displacing the frozen text embeddings along
the drift direction and upsampling to the visual feature resolution:
\begin{equation}
    T^{\ell,\pm} = \mathrm{Norm}\left(\mathbf{t}^{\pm} + \triangle^{\ell,\pm} \cdot \rho\right),
    \label{eq:drift}
\end{equation}
where each spatial position of $T^{\ell,\pm} \in \mathbb{R}^{D \times H_s \times W_s}$
holds a text representation uniquely adapted to the visual content at that location and depth.

\subsection{Drift-Guided Spatial Gating}

While VGTD generates spatially-adaptive text descriptors, the visual features $A^\ell$
still contain both normal and anomalous signals without distinction.
DGSG addresses this by using the drifted abnormal descriptor $T^{\ell,-}$ as a spatial
probe to identify and selectively enhance anomaly-relevant regions in $A^\ell$, producing
refined visual features $\tilde{A}^\ell$ that are better aligned with the drifted text
space for precise anomaly localization.
Both $A^\ell$ and $T^{\ell,-}$ are projected and reshaped into $K$ heads, where pixel-wise cosine similarity is computed for each head with a learnable temperature $\tau_k$:
\begin{equation}
    G^\ell_k = \sigma\!\left(\frac{\langle A^\ell_k,\, T^{\ell,-}_k \rangle}{\tau_k}\right),
    \quad k = 1, \ldots, K.
    \label{eq:gate}
\end{equation}
The $K$ per-head gates are fused via a lightweight convolution into a unified attention gate
$G^\ell \in \mathbb{R}^{H_s \times W_s}$.
The gated and refined feature is then obtained via:
\begin{equation}
    \tilde{A}^\ell = \mathrm{Conv}([A^\ell \odot (1 + G^\ell),\, A^\ell]) + A^\ell,
    \label{eq:skip}
\end{equation}
where the residual term $1 + G^\ell$ selectively amplifies anomaly-affine regions while
the skip connection retains low-level spatial details.

\subsection{Multi-Branch Anomaly Scoring}

To leverage complementary anomaly evidence from text-guided and appearance-level sources,
DriftAD aggregates three branches in MBAS to produce the final anomaly map.

\vspace{-0.5ex}
\paragraph{Clean Branch.}
We align the scale-aware features $V^\ell$ with the frozen text embeddings
$\{\mathbf{t}^{+}, \mathbf{t}^{-}\}$ via cosine similarity $\langle \cdot,  \cdot \rangle$ to produce per-layer logits:
\begin{equation}
    M^\ell_c = \left[\langle V^\ell,\, \mathbf{t}^{+} \rangle,\;
    \langle V^\ell,\, \mathbf{t}^{-} \rangle\right],
    \label{eq:clean_score}
\end{equation}
which preserves the zero-shot generalization capability of CLIP.

\vspace{-0.5ex}
\paragraph{Drift Branch.}
The refined features $\tilde{A}^\ell$ are matched against the drifted descriptors
$T^{\ell,+}$ and $T^{\ell,-}$, yielding spatially-aware logits that capture
defect-specific anomaly signals:
\begin{equation}
    M^\ell_d = \left[\langle \tilde{A}^\ell,\, T^{\ell,+} \rangle,\;
    \langle \tilde{A}^\ell,\, T^{\ell,-} \rangle\right].
    \label{eq:drift_score}
\end{equation}

\vspace{-0.5ex}
\paragraph{Support Set Branch.}
The patch features of the support set are stored in a memory bank $R$ across all layers.
The localization map $M_s$ is obtained by measuring the distance between each query patch
and its nearest neighbor in $R$:
\begin{equation}
    M_s = \mathrm{Up}\!\left(\sum_{\ell=1}^{4}\!\left(1 - \max_{r \in R}
    \langle F^\ell,\, r \rangle\right)\right).
    \label{eq:support}
\end{equation}
where $\mathrm{Up}(\cdot)$ denotes bilinear upsampling. This branch is activated only at inference time.

\vspace{-0.5ex}
\paragraph{Branch Fusion.}
The clean and drift branches are fused per layer to obtain $M_p$, which is then combined
with $M_s$ to produce the final anomaly map $M = \mathcal{F}(M_c, M_d, M_s)$:
\begin{gather}
    M_p = \mathrm{Up}\!\left(\mathrm{Norm}\sum_{\ell=1}^{4}
    \mathrm{softmax}\!\left((1-\mu)\, M^\ell_c + \mu\, M^\ell_d\right)\right),
    \label{eq:mp} \\
    M = \gamma M_p + (1 - \gamma)\, M_s,
    \label{eq:final}
\end{gather}
where $\mu$ balances cross-category generalization with defect sensitivity, and $\gamma$
balances the prediction and support branches.

\subsection{Loss Function}
\label{sec:loss}

We adopt binary cross-entropy, focal, and dice losses following prior work~\cite{jeong2023winclip}, and introduce two losses tailored to the proposed VGTD and DGSG modules.
The total training objective is:
\begin{equation}
    \mathcal{L} = \mathcal{L}_{\text{cls}} + \lambda_1\,\mathcal{L}_{\text{seg}}
    + \lambda_2\,\mathcal{L}_{\text{gate}} + \lambda_3\,\mathcal{L}_{\text{drift}},
    \label{eq:total_loss}
\end{equation}
where $\lambda_1 = 1.0$, $\lambda_2 = 0.2$, and $\lambda_3 = 2.5$.

\vspace{-0.5ex}
\paragraph{Classification and Segmentation Losses.}
Image-level supervision is applied solely to the CLS token score
$s_1 = \mathrm{FC}(F_{\text{cls}})(\mathbf{t}^{-})^\top$, decoupling the classification
objective from the support branch which is only available at inference time:
\begin{equation}
    \mathcal{L}_{\text{cls}} = \mathcal{L}_{\text{BCE}}(s_1,\, y),
    \label{eq:cls_loss}
\end{equation}
where $y \in \{0, 1\}$ is the image-level ground-truth label.
Per-layer predictions $M_p^\ell$ of the fused clean and drift branches are supervised with focal and dice losses:
\begin{equation}
    \begin{split}
    \mathcal{L}_{\text{seg}} =\;
    & \sum_{\ell=1}^{4} \mathcal{L}_{\text{Focal}}([I - M^\ell_p,\, M^\ell_p],\, G) \\
    & + \sum_{\ell=1}^{4} \left(\mathcal{L}_{\text{Dice}}(M^\ell_p,\, G)
    + \mathcal{L}_{\text{Dice}}(I - M^\ell_p,\, I - G)\right),
    \end{split}
    \label{eq:seg_loss}
\end{equation}
where $I$ denotes an all-ones map, $G$ denotes the pixel-level ground-truth mask of the synthesized pseudo-anomaly, and $[\cdot,\cdot]$ denotes channel-wise concatenation.

\vspace{-0.5ex}
\paragraph{Gate Supervision Loss.}
To encourage each attention gate $G^\ell$ to precisely localize anomalous regions, we
apply BCE loss directly on each gate:
\begin{equation}
    \mathcal{L}_{\text{gate}} = \sum_{\ell=1}^{4} \mathcal{L}_{\text{BCE}}(G^\ell,\, G).
    \label{eq:gate_loss}
\end{equation}

\vspace{-0.5ex}
\paragraph{Drift Separation Loss.}
To prevent representational collapse where $T^{\ell,+}$ and $T^{\ell,-}$ converge
toward similar representations, we enforce that their cosine similarity does not exceed
that of the original frozen embeddings by a margin $\epsilon$:
\begin{equation}
    \mathcal{L}_{\text{drift}} = \sum_{\ell=1}^{4}
    \left\|\max\!\left(
    \langle T^{\ell,+},\, T^{\ell,-} \rangle - \left(\langle \mathbf{t}^{+},\,
    \mathbf{t}^{-} \rangle - \epsilon\right),\; 0\right)\right\|^2,
    \label{eq:drift_loss}
\end{equation}
where $\epsilon = 0.05$. Unlike prompt learning methods, $\mathcal{L}_{\text{drift}}$
directly regularizes the drifted text geometry, preserving CLIP's discriminative
structure while allowing spatially-adaptive displacement.

\vspace{-0.5ex}
\paragraph{Training and Inference.}
During training, the support branch and top-$k$ scoring are not used, and the model is optimized with the four losses above, with image-level supervision applied only to $s_1$.
At inference, the final image-level score fuses $s_1$ with local evidence:
\begin{equation}
    s = \gamma s_1 + (1 - \gamma)\,\mathrm{Mean}(\mathrm{Top}\text{-}k(M)).
    \label{eq:score}
\end{equation}

\begin{table*}[t]
  \centering
  \caption{Performance comparisons on MVTec-AD and VisA datasets. AUROC measures image-level
  anomaly detection and pAUROC evaluates pixel-level anomaly localization. \textbf{Bold} and
  \underline{Underlined} indicate the best and second-best results respectively.}
  \vspace{-0.1in}
  \label{table:1}
  \setlength{\tabcolsep}{1.5mm}
  \resizebox{0.92\linewidth}{!}{
    \begin{tabular}{l ccc|ccc|ccc}
      \toprule
      \multirow{2}{*}{Method} &
      \multicolumn{3}{c|}{MVTec-AD (AUROC, pAUROC)} &
      \multicolumn{3}{c|}{VisA (AUROC, pAUROC)} &
      \multicolumn{3}{c}{Average (AUROC, pAUROC)} \\
      \cmidrule(lr){2-4} \cmidrule(lr){5-7} \cmidrule(lr){8-10}
      & 1-shot & 2-shot & 4-shot & 1-shot & 2-shot & 4-shot & 1-shot & 2-shot & 4-shot \\
      \midrule
      PatchCore~\cite{roth2022towards} {\scriptsize CVPR'22}
        & (83.4, 92.0) & (86.3, 93.3) & (88.8, 94.3)
        & (79.9, 95.4) & (81.6, 96.1) & (85.3, 96.8)
        & (81.7, 93.7) & (84.0, 94.7) & (87.1, 95.6) \\
      WinCLIP~\cite{jeong2023winclip} {\scriptsize CVPR'23}
        & (93.1, 95.2) & (94.4, 96.0) & (95.2, 96.2)
        & (83.8, 96.4) & (84.6, 96.8) & (87.3, 97.2)
        & (88.5, 95.8) & (89.5, 96.4) & (91.3, 96.7) \\
      AnomalyGPT~\cite{gu2024anomalygpt} {\scriptsize AAAI'24}
        & (94.1, 95.3) & (95.5, 95.6) & (96.3, 96.2)
        & (87.4, 96.2) & (88.6, 96.4) & (90.6, 96.7)
        & (90.8, 95.8) & (92.1, 96.0) & (93.5, 96.5) \\
      PromptAD~\cite{li2024promptad} {\scriptsize CVPR'24}
        & (94.6, 95.9) & (95.7, 96.2) & (96.6, 96.5)
        & (86.9, 96.7) & (88.3, 97.1) & (89.1, 97.4)
        & (90.8, 96.3) & (92.0, 96.7) & (92.9, 97.0) \\
      ResAD~\cite{yao2024resad} {\scriptsize NeurIPS'24}
        & (84.8, 93.4) & (87.2, 94.8) & (90.7, 95.8)
        & (80.9, 95.9) & (86.6, 96.5) & (89.3, 96.8)
        & (82.9, 94.7) & (86.9, 95.7) & (90.0, 96.3) \\
      KAG-prompt~\cite{tao2025kernel} {\scriptsize AAAI'25}
        & (95.8, 96.2) & (\underline{96.6}, 96.5) & (97.1, 96.7)
        & (\underline{91.6}, 97.0) & (92.7, 97.4) & (93.3, 97.7)
        & (\underline{93.7}, 96.6) & (\underline{94.7}, 97.0) & (95.2, 97.2) \\
      IIPAD~\cite{lv2025one} {\scriptsize ICLR'25}
        & (94.2, \underline{96.4}) & (95.7, 96.7) & (96.1, \underline{97.0})
        & (85.4, 96.9) & (86.7, 97.2) & (88.3, 97.4)
        & (89.8, 96.7) & (91.2, 97.0) & (92.2, 97.2) \\
      FiLo++~\cite{gu2026filo++} {\scriptsize TCSVT'26}
        & (95.0, 96.2) & (95.8, \underline{96.9}) & (96.3, 96.6)
        & (88.3, \underline{97.3}) & (88.6, 97.5) & (89.8, \underline{97.9})
        & (91.7, \underline{96.8}) & (92.2, 97.2) & (93.1, \underline{97.3}) \\
      FocusPatch-AD~\cite{ding2025focuspatch} {\scriptsize TIP'26}
        & (\underline{96.0}, \underline{96.4}) & (96.4, \underline{96.9}) & (\underline{97.4}, \textbf{97.2})
        & (91.0, 96.8) & (\underline{92.8}, \underline{97.7}) & (\underline{93.6}, \underline{97.9})
        & (93.5, 96.6) & (94.6, \underline{97.3}) & (\underline{95.5}, \textbf{97.6}) \\
      \midrule
      \textbf{DriftAD (Ours)}
        & (\textbf{97.2}, \textbf{96.8}) & (\textbf{97.7}, \textbf{97.0}) & (\textbf{98.0}, \textbf{97.2})
        & (\textbf{93.1}, \textbf{97.4}) & (\textbf{93.3}, \textbf{97.8}) & (\textbf{94.0}, \textbf{98.0})
        & (\textbf{95.2}, \textbf{97.1}) & (\textbf{95.5}, \textbf{97.4}) & (\textbf{96.0}, \textbf{97.6}) \\
      \bottomrule
    \end{tabular}
  }
  \vspace{-0.1in}
\end{table*}
\begin{table*}[t]
  \centering
  \caption{Performance comparisons on MVTec-AD and VisA datasets. AUPR measures the area
  under the precision-recall curve for anomaly detection, and PRO evaluates pixel-level
  anomaly localization by computing the per-region overlap between predicted and
  ground-truth anomaly regions. \textbf{Bold} and \underline{Underlined} indicate the best
  and second-best results respectively.}
  \vspace{-0.1in}
  \label{table:2}
  \setlength{\tabcolsep}{1.5mm}
  \resizebox{0.92\linewidth}{!}{
    \begin{tabular}{l ccc|ccc|ccc}
      \toprule
      \multirow{2}{*}{Method} &
      \multicolumn{3}{c|}{MVTec-AD (AUPR, PRO)} &
      \multicolumn{3}{c|}{VisA (AUPR, PRO)} &
      \multicolumn{3}{c}{Average (AUPR, PRO)} \\
      \cmidrule(lr){2-4} \cmidrule(lr){5-7} \cmidrule(lr){8-10}
      & 1-shot & 2-shot & 4-shot & 1-shot & 2-shot & 4-shot & 1-shot & 2-shot & 4-shot \\
      \midrule
      PatchCore~\cite{roth2022towards} {\scriptsize CVPR'22}
        & (92.2, 79.7) & (93.8, 82.3) & (94.5, 84.3)
        & (82.8, 80.5) & (84.8, 82.6) & (87.5, 84.9)
        & (87.5, 80.1) & (89.3, 82.4) & (91.0, 84.6) \\
      WinCLIP~\cite{jeong2023winclip} {\scriptsize CVPR'23}
        & (96.5, 87.1) & (97.0, 88.4) & (97.3, 89.0)
        & (85.1, 85.1) & (85.8, 86.2) & (88.8, 87.6)
        & (90.8, 86.1) & (91.4, 87.3) & (93.1, 88.3) \\
      AnomalyGPT~\cite{gu2024anomalygpt} {\scriptsize AAAI'24}
        & (95.9, 89.5) & (96.8, 90.0) & (97.6, 90.7)
        & (88.7, 82.9) & (89.0, 83.4) & (91.3, 84.6)
        & (92.3, 86.2) & (92.9, 86.7) & (94.4, 87.7) \\
      PromptAD~\cite{li2024promptad} {\scriptsize CVPR'24}
        & (97.1, 87.9) & (97.9, 88.5) & (98.5, 90.5)
        & (88.4, 85.1) & (90.0, 85.8) & (90.8, 86.2)
        & (92.8, 86.5) & (94.0, 87.2) & (94.7, 88.4) \\
      ResAD~\cite{yao2024resad} {\scriptsize NeurIPS'24}
        & (92.7, 83.3) & (93.9, 85.5) & (95.7, 88.7)
        & (83.7, 79.6) & (88.3, 82.3) & (90.7, 84.1)
        & (88.2, 81.4) & (91.1, 83.9) & (93.2, 86.4) \\
      KAG-prompt~\cite{tao2025kernel} {\scriptsize AAAI'25}
        & (\underline{98.1}, \underline{90.8}) & (\underline{98.5}, \underline{91.1}) & (\underline{98.8}, \underline{91.4})
        & (\underline{93.2}, 85.2) & (\underline{94.2}, 86.7) & (\underline{94.6}, 87.6)
        & (\underline{95.7}, 88.0) & (\underline{96.4}, 88.9) & (\underline{96.7}, 89.5) \\
      IIPAD~\cite{lv2025one} {\scriptsize ICLR'25}
        & (97.2, 89.8) & (97.9, 90.3) & (98.1, 91.2)
        & (87.5, \textbf{87.3}) & (88.6, \textbf{87.9}) & (89.6, \textbf{88.3})
        & (92.4, \underline{88.6}) & (93.3, \underline{89.1}) & (93.9, \underline{89.8}) \\
      FocusPatch-AD~\cite{ding2025focuspatch} {\scriptsize TIP'26}
        & (97.3, 88.9) & (97.8, 89.1) & (98.6, 89.3)
        & (88.2, 80.3) & (89.8, 86.1) & (90.8, 86.6)
        & (92.8, 84.6) & (93.8, 87.6) & (94.7, 87.9) \\
      \midrule
      \textbf{DriftAD (Ours)}
        & (\textbf{98.8}, \textbf{92.2}) & (\textbf{99.0}, \textbf{92.6}) & (\textbf{99.2}, \textbf{92.7})
        & (\textbf{94.8}, \underline{86.7}) & (\textbf{95.0}, \underline{87.8}) & (\textbf{95.3}, \textbf{88.3})
        & (\textbf{96.8}, \textbf{89.5}) & (\textbf{97.0}, \textbf{90.2}) & (\textbf{97.3}, \textbf{90.5}) \\
      \bottomrule
    \end{tabular}
  }
  \vspace{-0.1in}
\end{table*}
\begin{table}[t]
\centering
\caption{Performance comparison of AUROC and pAUROC on MVTec-AD and VisA with unified many-shot anomaly detection methods.}
\label{tab:3}
  \vspace{-0.1in}
\resizebox{0.8\linewidth}{!}{
\begin{tabular}{lc|cc}
\toprule
Method & Setting & MVTec-AD & VisA  \\ \midrule
RegAD~\cite{huang2022registration} {\scriptsize ECCV'22} & 8-shot & (91.2, 96.7) & -- \\
FastRecon~\cite{fang2023fastrecon} {\scriptsize ICCV'23} & 8-shot & (95.2, 97.3) & -- \\
DeSTSeg~\cite{zhang2023destseg} {\scriptsize CVPR'23} & full-shot & (98.6, 97.4) & --\\
UniAD~\cite{you2022unified} {\scriptsize NeurIPS'22} & full-shot & (96.5, 96.8) & (91.9, 98.6) \\
OmniAL~\cite{zhao2023omnial} {\scriptsize CVPR'23}  & full-shot  & (97.2, 98.3) & (87.8, 96.6) \\
HVQ-Trans~\cite{lu2023hierarchical} {\scriptsize NeurIPS'23} & full-shot  & (98.0, 97.3) & (93.2, 98.7) \\
DiAD~\cite{he2024diffusion} {\scriptsize AAAI'24}  & full-shot  & (97.2, 96.8) & (86.8, 96.0) \\ 
\midrule
\multirow{3}{*}{\textbf{DriftAD (Ours)}} & 1-shot & (97.2, 96.8) & (93.1, 97.4) \\
 & 2-shot & (97.7, 97.0) & (93.3, 97.8) \\
 & 4-shot & (98.0, 97.2) & (94.0, 98.0) \\ 
\bottomrule
\end{tabular}
}
\vspace{-0.1in}
\end{table}
\section{Experiments}
\subsection{Experimental Settings}

We conduct experiments on MVTec-AD~\cite{bergmann2019mvtec} and VisA~\cite{zou2022spot}.
MVTec-AD contains 5,354 high-resolution images across 15 categories (5 textures and 10
objects), with 3,629 normal training images and 1,725 test images containing both normal and
anomalous samples.
VisA contains 10,821 images across 12 subsets, of which 9,621 are normal and 1,200 are
anomalous.
Following the standard few-shot anomaly detection protocol, training sets contain only normal
samples, while test sets contain both normal and anomalous samples.
We synthesize anomalous images from normal training samples using
NSA~\cite{schluter2022natural} to provide pixel-level supervision during training, and
following AnomalyGPT~\cite{gu2024anomalygpt}, we train on one dataset and perform few-shot
testing on the other to avoid data leakage.
We use the area under the ROC curve (AUROC) and average precision (AUPR) for image-level
anomaly detection, and pixel-wise AUROC (pAUROC) and per-region overlap (PRO) for
pixel-level anomaly localization.
 
All images are resized to $224 \times 224$.
We use the frozen OpenCLIP ViT-H/14 image and text encoders released with
ImageBind-H~\cite{girdhar2023imagebind}. Although implemented through the ImageBind
checkpoint, DriftAD does not use its additional modality encoders. Visual features are
extracted from layers 8, 16, 24, and 36 of the vision encoder, yielding four hierarchical
feature maps at $H_s = W_s = 16$.
For the Visually-Guided Text Drift module, the layer-specific grid sizes are set to
$g_\ell \in \{4, 8, 8, 16\}$ from shallow to deep layers, with drift scale $\rho$ clamped to
$[0.01, 0.2]$ and margin $\epsilon = 0.05$.
The number of heads in the similarity gating is $K = 4$.
The fusion coefficient between the clean and drift branches is set to $\mu = 0.7$.
Following~\cite{tao2025kernel}, for few-shot inference, the fusion weight is set to $\gamma = 0.1$ for both the anomaly map and image-level score, and the top-$k$ strategy uses $k = 30$.
All models are optimized with Adam with $\beta_1 = 0.9$, $\beta_2 = 0.95$, and weight decay
$0.001$, using a warmup decay learning rate scheduler with warmup steps of 200, learning
rate $1 \times 10^{-3}$, and batch size 16 on two RTX-3090 GPUs, trained for 50 epochs
on MVTec-AD and 80 epochs on VisA.

\begin{figure*}[t]
    \centering
    \includegraphics[width=0.48\textwidth]{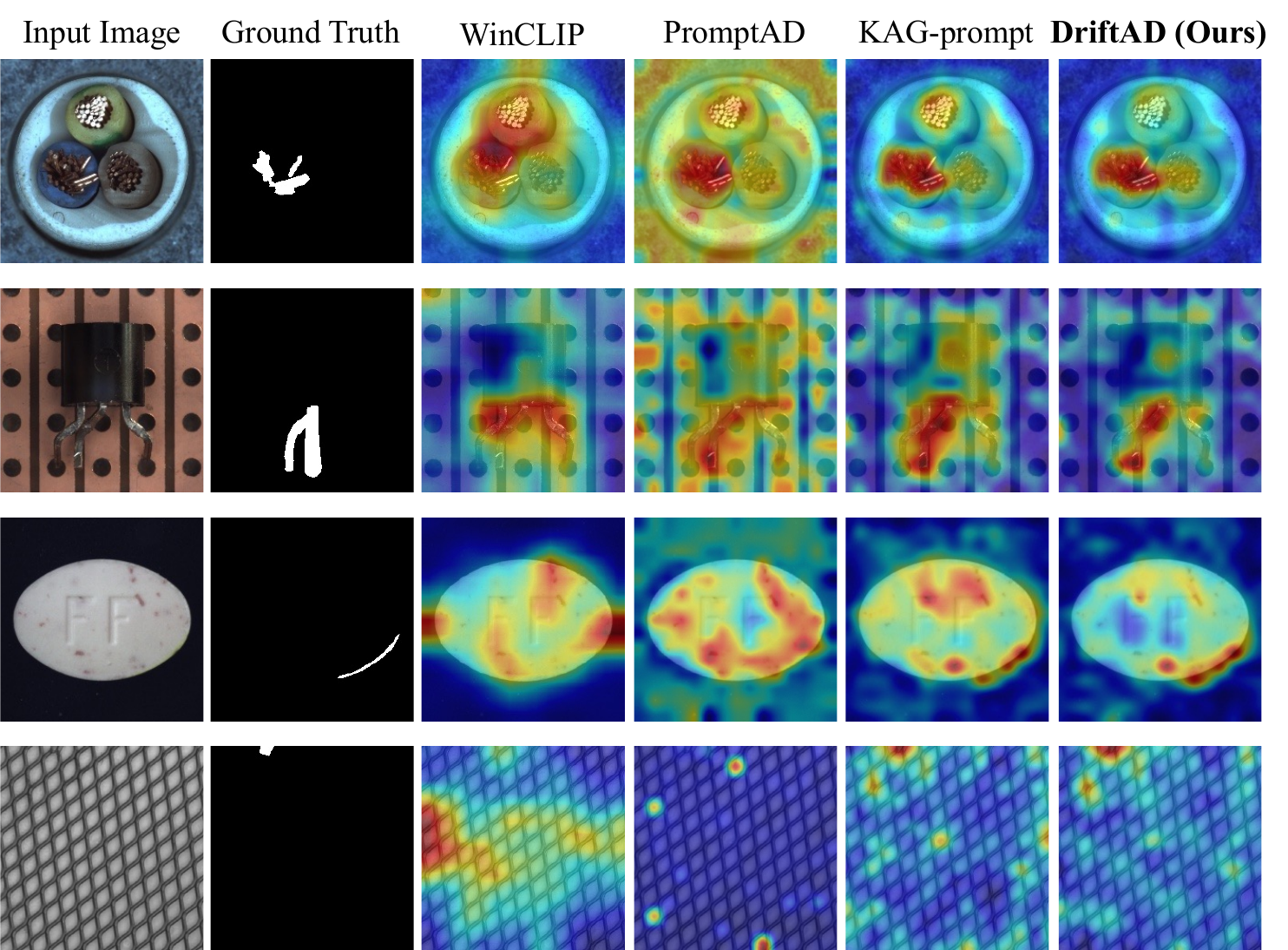}
    \hspace{0.001\textwidth} 
    \includegraphics[width=0.48\textwidth]{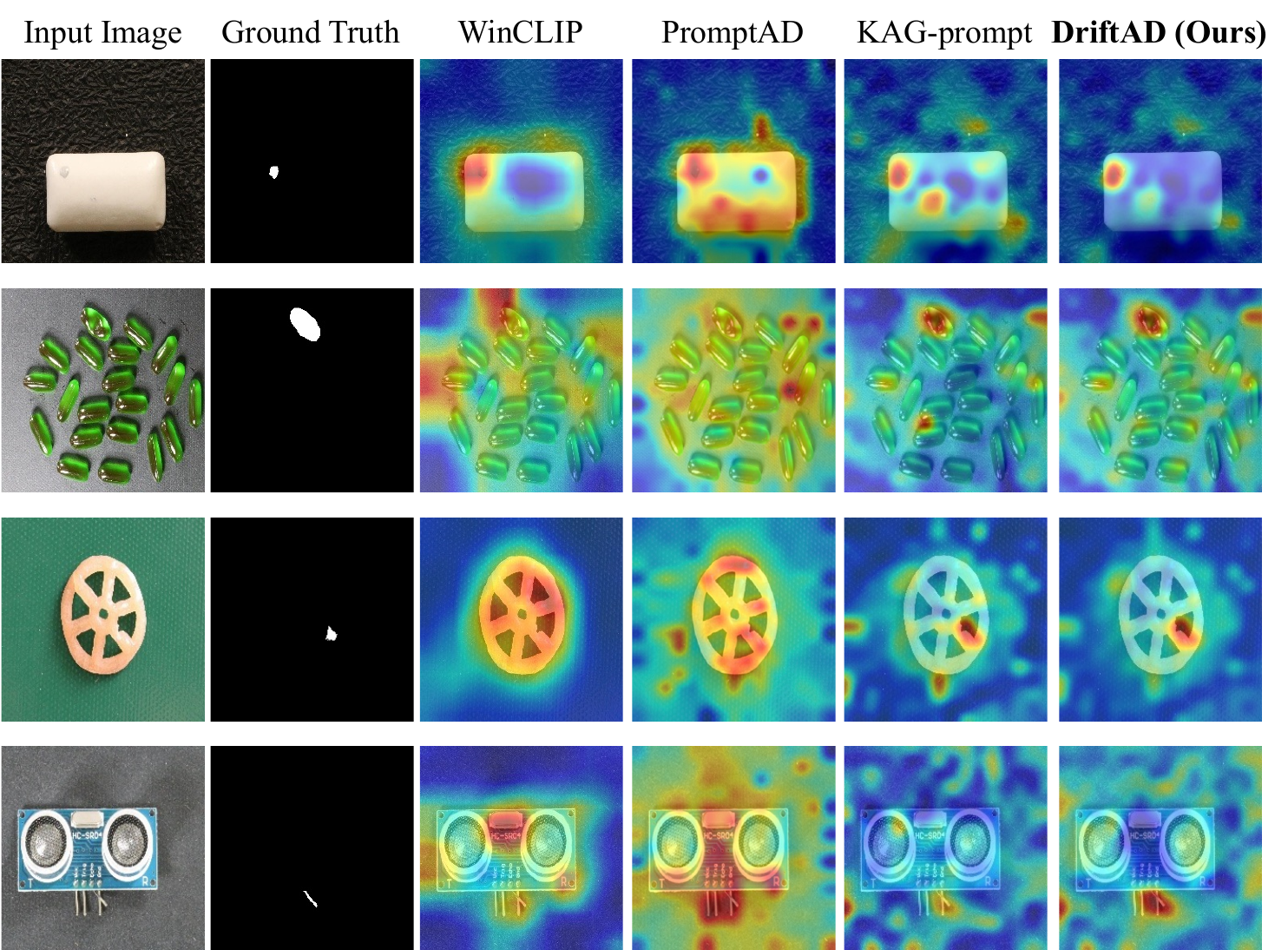}
    
    \vspace{-0.1in}
   \caption{Qualitative comparison of anomaly localization on MVTec-AD (left) and VisA (right). DriftAD yields clearer anomaly maps with reduced background interference compared to the baseline methods.}
    \vspace{-0.1in}
    \label{fig:vis}
\end{figure*}

\begin{figure}[t]
    \centering
\includegraphics[width=0.94\columnwidth]{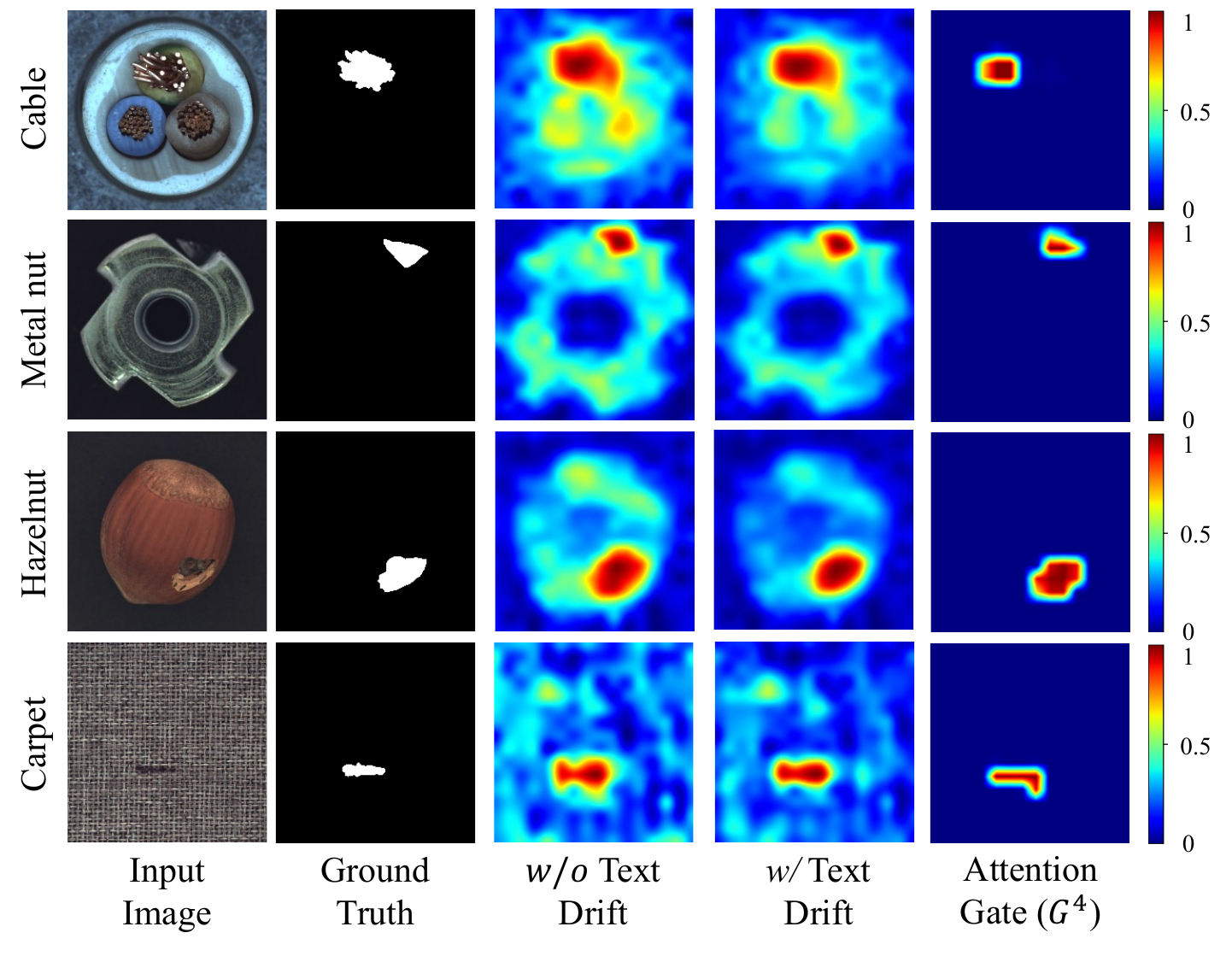}
    \vspace{-0.1in}
    \caption{Visualization of anomaly maps and the layer-4 attention gate $G^4$ on MVTec-AD. Without text drift, the anomaly map produces diffuse activations across large regions. With text drift enabled, the anomaly map becomes sharper, and the attention gate $G^4$ confirms that the drifted abnormal descriptor precisely localizes defect regions.}
    \vspace{-0.1in}
    \label{fig:vis_baseline}
\end{figure}

We compare DriftAD with recent state-of-the-art FSAD methods, including
PatchCore~\cite{roth2022towards},
WinCLIP~\cite{jeong2023winclip}, AnomalyGPT~\cite{gu2024anomalygpt},
PromptAD~\cite{li2024promptad}, ResAD~\cite{yao2024resad}, KAG-prompt~\cite{tao2025kernel},
FiLo++~\cite{gu2026filo++}, and FocusPatch-AD~\cite{ding2025focuspatch}.
In particular, AnomalyGPT, KAG-prompt, and DriftAD use the same frozen OpenCLIP
ViT-H/14 visual backbone released with ImageBind-H, enabling a backbone-matched comparison.
Unlike existing approaches that rely on static text embeddings, DriftAD
dynamically conditions its text descriptors on local visual context at each encoder scale,
enabling precise text-patch alignment while preserving the zero-shot generalization of CLIP.
As presented in Table~\ref{table:1}, DriftAD achieves the best average performance across
all 1-, 2-, and 4-shot settings on both datasets, improving average AUROC by up to
$1.5\%$ and average pAUROC by up to $0.5\%$ over the best competing method KAG-prompt~\cite{tao2025kernel} in the 1-shot setting.
The gains are most consistent on VisA, where the higher spatial diversity of defects
better reveals the advantage of spatially-adaptive text drift.
Table~\ref{table:2} further shows average improvements of up to $1.1\%$ in AUPR and
$1.5\%$ in PRO over KAG-prompt in the 1-shot setting, confirming that text drift yields consistent pixel-level localization
gains across both datasets.

\begin{figure}[t]
    \centering
\includegraphics[width=0.89\columnwidth]{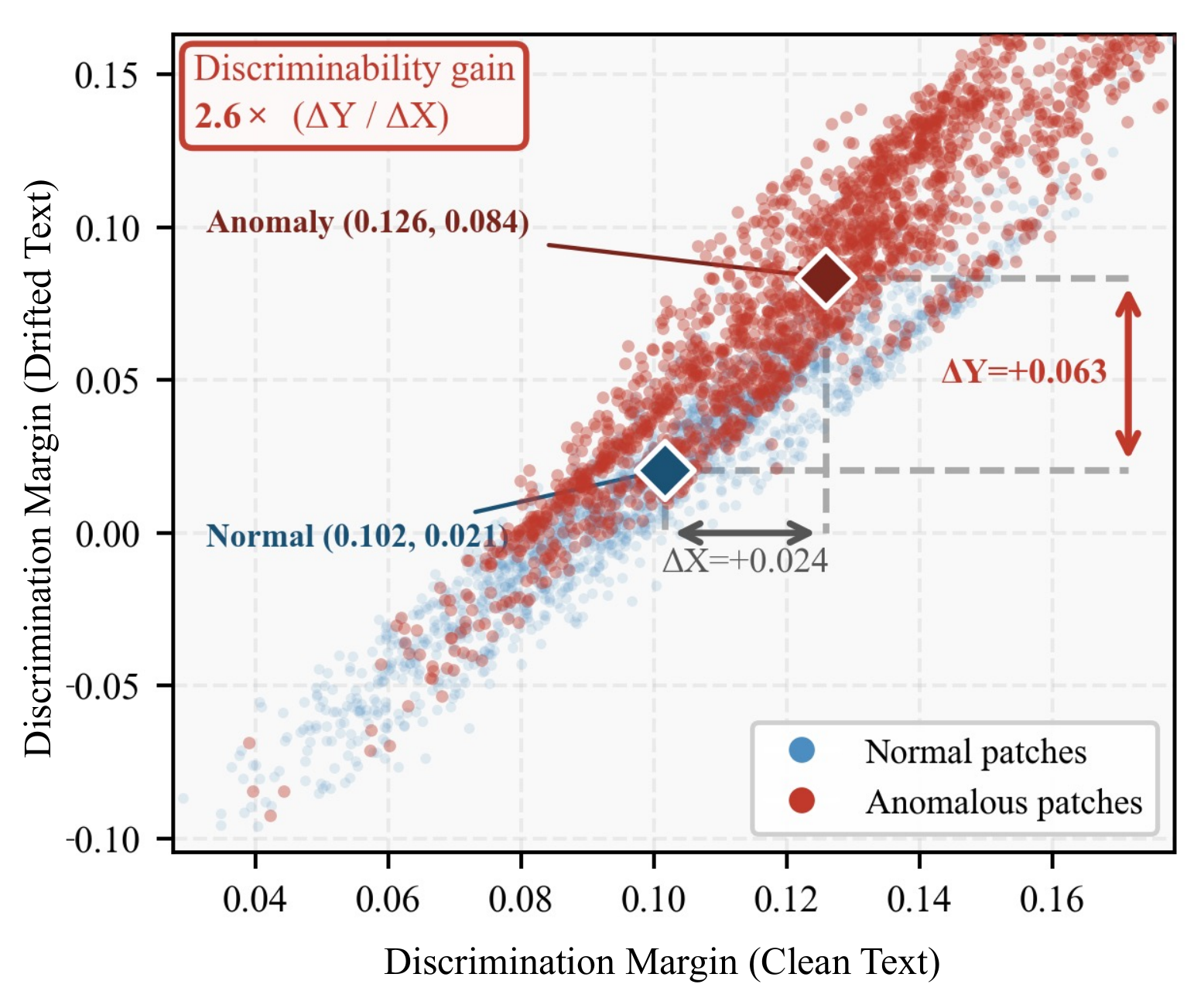}
    \vspace{-0.1in}
    \caption{Patch-level discrimination margin comparing clean and drifted text on MVTec-AD features. While clean text assigns similar scores to normal and anomalous patches, drifted text achieves a $\mathbf{2.6\times}$ larger gap by adapting to visual content, confirming the advantage of layer-wise, spatially-adaptive descriptors over global static text descriptors.}
    \vspace{-0.15in}
    \label{fig:vis_text}
\end{figure}

We further compare DriftAD with representative 8-shot and full-shot anomaly detection methods. RegAD~\cite{huang2022registration} and FastRecon~\cite{fang2023fastrecon} are evaluated under the 8-shot setting, while DeSTSeg~\cite{zhang2023destseg}, UniAD~\cite{you2022unified}, OmniAL~\cite{zhao2023omnial}, HVQ-Trans~\cite{lu2023hierarchical}, and DiAD~\cite{he2024diffusion} are full-shot baselines. As shown in Table~\ref{tab:3}, DriftAD with only 4-shot support achieves competitive or superior performance against these baselines. Notably, DriftAD substantially outperforms the 8-shot methods and surpasses full-shot DiAD by $0.8\%$ AUROC and $0.4\%$ pAUROC on MVTec-AD, while remaining competitive with the full-shot methods on VisA. These results demonstrate strong anomaly detection and localization performance with minimal normal support.
 

\subsection{Qualitative Comparison}
Figure~\ref{fig:vis} presents qualitative comparisons on MVTec-AD and VisA under the
1-shot setting against WinCLIP~\cite{jeong2023winclip}, PromptAD~\cite{li2024promptad}, and KAG-prompt~\cite{tao2025kernel}.
WinCLIP relies on fixed hand-crafted templates without any learned adaptation, producing
coarse activation maps that highlight broad regions rather than precise defect boundaries.
PromptAD introduces learnable prompt tokens to improve text-visual alignment, but the
learned tokens remain a single global descriptor shared across all spatial positions and
encoder depths, causing the model to miss spatially localized defects such as thin
scratches and point anomalies.
KAG-prompt strengthens the visual side through graph-based cross-layer feature interaction,
improving feature discriminability, but its text descriptors remain globally fixed
throughout, limiting the model's ability to adapt text-visual alignment to individual
defect locations.
DriftAD addresses this fundamental limitation from the text side: by conditioning the drift
on local visual context at each encoder depth, each spatial position receives a text
descriptor uniquely tailored to its visual content, enabling precise localization that
adapts to the shape, scale, and spatial distribution of each defect.
This advantage is particularly evident on VisA, where defects are small and scattered across structurally complex objects, highlighting the ability of our method to suppress background activations while preserving true anomaly signals, in contrast to static text descriptors.

\subsection{Visualization Analysis}

\paragraph{Effect of Text Drift and Spatial Gating.}
Figure~\ref{fig:vis_baseline} presents a side-by-side comparison of the anomaly map
without text drift, with text drift enabled, and the layer-4 attention gate $G^4$ on
representative MVTec-AD samples across four categories.
Without text drift, the anomaly map produces diffuse activations that broadly highlight
large regions, as the clean branch relies solely on static global text embeddings.
With text drift enabled, VGTD generates layer-wise, spatially-adaptive descriptors $T^{\ell,\pm}$
conditioned on local visual context, and DGSG uses $T^{\ell,-}$ as a spatial probe to
compute per-layer attention gates $\{G^\ell\}$ that selectively enhance anomaly-relevant
regions.
To explore the role of spatial gating, we visualize the layer-4 gate $G^4$, which
reveals that the drifted descriptor precisely highlights defect regions while suppressing
background responses, explaining the sharper anomaly maps produced by the full model.

\begin{table}[t]
\centering
\caption{Ablation study of architecture modules on MVTec-AD and VisA (1-shot).}
\label{tab:ablation_modules}
\vspace{-0.1in}
\resizebox{0.98\columnwidth}{!}{%
\begin{tabular}{lccccc|cc}
\toprule
\multirow{2}{*}{Variant} 
& \multirow{2}{*}{ASA} 
& \multirow{2}{*}{VGTD} 
& \multirow{2}{*}{DGSG} 
& \multicolumn{2}{c|}{MVTec-AD} 
& \multicolumn{2}{c}{VisA} \\
\cmidrule(lr){5-6}\cmidrule(lr){7-8}
& & & & AUROC & pAUROC & AUROC & pAUROC \\
\midrule
Baseline &            &            &            & 95.4 & 95.9 & 91.5 & 97.0 \\
+ ASA    & \checkmark &            &            & 95.6 & 95.9 & 91.6 & 97.0 \\
+ VGTD   & \checkmark & \checkmark &            & 95.7 & 95.9 & 91.7 & 97.0 \\
+ DGSG   & \checkmark & \checkmark & \checkmark & \textbf{96.8} & \textbf{96.7} & \textbf{92.3} & \textbf{97.2} \\
\bottomrule
\end{tabular}%
}
\vspace{-0.1in}
\end{table}

\begin{table}[t]
\centering
\caption{Ablation study of loss functions on MVTec-AD and VisA (1-shot).}
\vspace{-0.1in}
\label{tab:ablation_losses}
\resizebox{0.98\columnwidth}{!}{%
\begin{tabular}{lcc cc|cc}
\toprule
\multirow{2}{*}{Variant} 
& \multirow{2}{*}{$\mathcal{L}_{\text{gate}}$} 
& \multirow{2}{*}{$\mathcal{L}_{\text{drift}}$} 
& \multicolumn{2}{c|}{MVTec-AD} 
& \multicolumn{2}{c}{VisA} \\
\cmidrule(lr){4-5}\cmidrule(lr){6-7}
& & & AUROC & pAUROC & AUROC & pAUROC \\
\midrule
Base Architecture                     &            &            & 96.8 & 96.7 & 92.3 & 97.2 \\
+ $\mathcal{L}_{\text{gate}}$  & \checkmark &            & 97.0 & 96.7 & 92.6 & 97.2 \\
Full Model (Ours)                    & \checkmark & \checkmark & \textbf{97.2} & \textbf{96.8} & \textbf{93.1} & \textbf{97.4} \\
\bottomrule
\end{tabular}%
}
\vspace{-0.1in}
\end{table}

\begin{table}[t]
\centering
\caption{Ablation study on layer-specific grid sizes $\{g_1, g_2, g_3, g_4\}$ on MVTec-AD (1-shot).}
\label{tab:ablation_grid}
\vspace{-0.1in}
\resizebox{0.9\columnwidth}{!}{%
\begin{tabular}{l|cccc|cc}
\toprule
Configuration & $g_1$ & $g_2$ & $g_3$ & $g_4$ & AUROC & pAUROC \\
\midrule
Uniform-4    & 4  & 4  & 4  & 4  & 97.2 & 96.6 \\
Uniform-8    & 8  & 8  & 8  & 8  & 97.1 & 96.7 \\
Uniform-16   & 16 & 16 & 16 & 16 & 96.3 &  96.2 \\
Descending   & 16 & 8  & 8  & 4  & 96.7 & 96.5 \\
Ascending (Ours)         & 4  & 8  & 8  & 16 & \textbf{97.2} & \textbf{96.8}\\
\bottomrule
\end{tabular}%
}
\vspace{-0.1in}
\end{table}

\vspace{-0.5ex}
\paragraph{Text Drift Discriminability.}
Figure~\ref{fig:vis_text} plots the patch-level discrimination margin, defined as the
cosine similarity difference between the abnormal and normal text descriptors, for each
patch on the MVTec-AD test set.
Each point represents a patch from a test image, colored by its ground-truth label
(normal or anomalous), with the $x$-axis showing the score under static clean text
and the $y$-axis showing the score under drifted text on the same visual features.
A larger gap between the two class distributions along an axis indicates stronger
discriminability.
While clean text yields a gap of $\Delta X = +0.024$, drifted text achieves a
$\mathbf{2.6\times}$ larger gap of $\Delta Y = +0.063$, purely from adapting the
text descriptors to local visual content without modifying the visual features.
This confirms that layer-wise, spatially-adaptive descriptors possess significantly greater discriminative power in distinguishing anomalous patches from normal ones, compared to static global embeddings.

\subsection{Ablation Study}
\paragraph{Effect of Proposed Components.}
Table~\ref{tab:ablation_modules} ablates the contribution of each module by progressively adding components to a clean-branch-only baseline. Adding Anomaly Signal Amplification (ASA) yields modest improvements in AUROC, as its spatial and frequency branches expose subtle defect signals otherwise suppressed by dominant normal patterns. Adding Visually-Guided Text Drift (VGTD) alone produces only marginal pAUROC gains. Although $T^{\ell,\pm}$ provide layer-wise, spatially-adaptive descriptors in the CLIP text space, the corresponding visual features $A^\ell$ remain unchanged in the original feature space, creating a residual text-visual mismatch that limits the benefit of text drift. Drift-Guided Spatial Gating (DGSG) addresses this mismatch by using $T^{\ell,-}$ as a spatial probe to refine $A^\ell$ into $\tilde{A}^\ell$, selectively emphasizing regions consistent with the drifted anomaly descriptors and improving text-visual correspondence. Together, VGTD and DGSG improve the ASA-only baseline by $+1.1\%$ AUROC and $+0.8\%$ pAUROC on MVTec-AD, and $+0.6\%$ AUROC and $+0.2\%$ pAUROC on VisA, confirming that explicit alignment with the drifted text space is essential to fully exploit the drift mechanism.

\vspace{-0.5ex}
\paragraph{Effect of Proposed Losses.}
Table~\ref{tab:ablation_losses} ablates the two losses proposed for our modules, where
the base architecture uses only the standard classification and segmentation losses, i.e., $\mathcal{L}_{\text{cls}}$ and $\mathcal{L}_{\text{seg}}$. 
Specifically, $\mathcal{L}_{\text{gate}}$ is tailored for DGSG, supervising each per-layer attention gate $G^\ell$ with ground-truth masks to enforce spatially discriminative gating, while $\mathcal{L}_{\text{drift}}$ is designed for VGTD to regularize $T^{\ell,+}$ and $T^{\ell,-}$, preventing them from converging toward similar representations and preserving inter-class separability in the CLIP embedding space.
Adding $\mathcal{L}_{\text{gate}}$ improves performance over the base model, and further incorporating $\mathcal{L}_{\text{drift}}$ yields additional gains. The full model achieves the best results on both MVTec-AD and VisA, demonstrating the complementarity of the two losses.

\vspace{-0.5ex}
\paragraph{Effect of Layer-Specific Grid Size.}
Table~\ref{tab:ablation_grid} examines different grid configurations for the VGTD module.
Uniform-16 performs worst, as assigning a fine-grained $16{\times}16$ grid to all layers
forces shallow layers to generate overly detailed drift fields beyond their semantic capacity.
The Descending configuration underperforms our design, confirming that deeper layers benefit
from finer-grained drift fields to exploit their richer spatial semantics.
Our layer-adaptive configuration $\{4, 8, 8, 16\}$ achieves the best pAUROC of $96.8\%$,
validating that matching grid resolution to encoder depth is key to effective
spatially-adaptive text drift.

\section{Conclusion}
We present DriftAD, a few-shot anomaly detection framework that addresses a key limitation of existing VLM-based methods: globally static text representations that cannot adapt to the spatial distribution and visual characteristics of defects. DriftAD first employs an Anomaly Signal Amplification module to highlight subtle defect cues. Guided by these visual contexts, our Visually-Guided Text Drift transforms frozen CLIP embeddings into layer-wise, spatially-adaptive anomaly descriptors. These descriptors then act as spatial probes in a Drift-Guided Spatial Gating mechanism to enhance anomaly-relevant visual features. Finally, predictions from the drifted descriptors are fused with a clean text branch, balancing defect-specific sensitivity with CLIP's generalization ability. Extensive experiments on MVTec-AD and VisA demonstrate consistent state-of-the-art performance.

\begin{acks}
This research is supported by the National Research Foundation, Prime Minister’s Office, Singapore, and the Ministry of Digital Development and Information, under its Online Trust and Safety (OTS) Research Programme (MDDI-OTS-001). Any opinions, findings and conclusions or recommendations expressed in this material are those of the author(s) and do not reflect the views of National Research Foundation, Prime Minister’s Office, Singapore, the Ministry of Digital Development and Information, or the Centre for Advanced Technologies in Online Safety.
\end{acks}

\bibliographystyle{ACM-Reference-Format}
\balance
\bibliography{sample-base}

\clearpage
\appendix
\twocolumn[{
\centering
{\LARGE\bfseries Supplementary Material\par}
\vspace{0.5em}
{\Large\bfseries DriftAD: Visually-Guided Text Drift for Few-Shot Industrial Anomaly Detection\par}
\vspace{3em}
}]

\section{Static Text Prompt Construction}
\label{sec:prompt}

Following standard practice in VLM-based anomaly
detection~\cite{jeong2023winclip,tao2025kernel}, normal and abnormal
text embeddings $\mathbf{t}^{+}$ and $\mathbf{t}^{-}$ are obtained via
prompt ensemble over a set of hand-crafted templates.
Each prompt is composed of a state-level descriptor [state] (denoted $s$)
that characterizes the condition of the object, and a template-level
sentence that wraps $s$ with the category name [class] (denoted $c$).

\vspace{-0.5ex}
\paragraph{State-level descriptors.}
For the normal class, descriptors include ``flawless [c]'', ``perfect [c]'',
``unblemished [c]'', ``[c] without flaw'', and ``[c] without defect''.
For the abnormal class, descriptors include ``damaged [c]'', ``broken [c]'',
``[c] with flaw'', ``[c] with defect'', and ``[c] with damage'', where [c]
is substituted with the category name at inference time.

\vspace{-0.5ex}
\paragraph{Template-level sentences.}
Each state descriptor is embedded into a sentence template
following~\cite{tao2025kernel,ding2025focuspatch}, using ``a photo of a
[s]'' and ``a photo of the [s]'' as the basic templates. Additional
templates such as ``a cropped photo of the [s]'', ``a close-up photo of a
[s]'', and ``a black and white photo of the [s]'' are included to improve
coverage.

\vspace{-0.5ex}
\paragraph{Prompt ensemble.}
All prompts from each class are passed through the frozen CLIP text encoder
and averaged within each class to produce $\mathbf{t}^{+}$ and
$\mathbf{t}^{-} \in \mathbb{R}^{D}$.
These embeddings are static and input-agnostic: computed once per category
and fixed throughout inference, they carry no information about the spatial
or structural characteristics of the query image.
Rather than using $\mathbf{t}^{+}$ and $\mathbf{t}^{-}$ directly for
text-visual matching, DriftAD treats them as base anchors for VGTD
(Figure~\ref{fig:motivation} of the main paper), which transforms them into layer-wise,
spatially-adaptive anomaly descriptors $T^{\ell,\pm}$.

\section{Anomaly synthesis for Training.}
Anomalous training samples are synthesized using
NSA~\cite{schluter2022natural}, which generates pseudo-anomalies by
blending patches from different normal images following the default
settings of the original implementation.
Pixel-level binary masks are automatically generated alongside each
synthesized sample and used as ground-truth supervision for
$\mathcal{L}_{\text{seg}}$ and $\mathcal{L}_{\text{gate}}$.
Text prompt construction follows Section~\ref{sec:prompt}, with the
resulting base embeddings $\mathbf{t}^{+}$ and $\mathbf{t}^{-}$ serving
as static anchors for VGTD throughout inference.

\begin{figure}[t]
    \centering
\includegraphics[width=0.92\columnwidth]{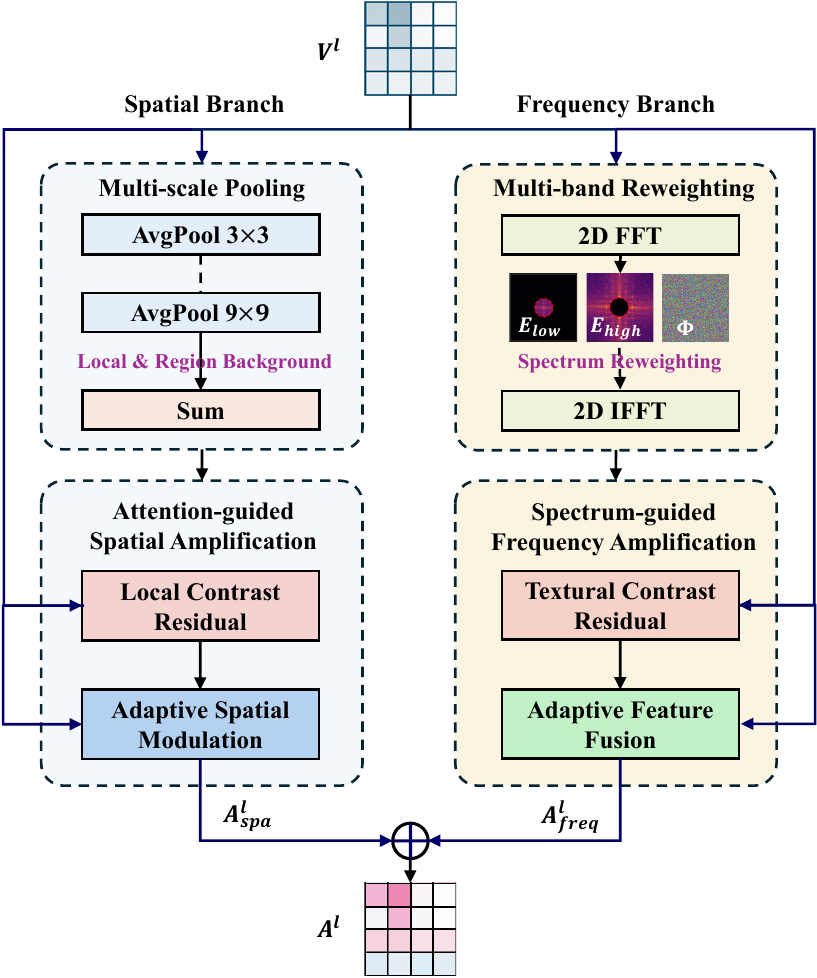}
\caption{Detailed architecture of the Anomaly Signal Amplification (ASA) module.
The spatial branch extracts multi-scale backgrounds via $3{\times}3$ and $9{\times}9$
average pooling and applies attention-guided modulation on the spatial residual.
The frequency branch decomposes $V^\ell$ via 2D FFT into low-frequency components
$\mathbf{E}_{\text{low}}$, high-frequency components $\mathbf{E}_{\text{high}}$,
and phase $\boldsymbol{\Phi}$, adaptively reweights the spectrum, and reconstructs
via IFFT to obtain the spectral residual.
The two branch outputs are fused to produce
amplified features $A^\ell$.}
\label{fig:asa}
\vspace{-0.1in}
\end{figure}

\section{Details of Anomaly Signal Amplification}

Figure~\ref{fig:asa} provides a detailed illustration of the Anomaly Signal Amplification (ASA) module introduced
in the corresponding subsection of the main paper.
ASA processes scale-aware features $V^\ell$ through two parallel branches to amplify
anomaly-relevant signals before text-visual matching.

\vspace{-0.5ex}
\paragraph{Spatial branch.}
As shown in the left panel of Figure~\ref{fig:asa}, the spatial branch begins with
Multi-scale Pooling, applying $3{\times}3$ and $9{\times}9$ average pooling to
obtain local and region-level background estimates of $V^\ell$.
The two smoothed outputs are each subtracted from $V^\ell$ and summed to form the
Local Contrast Residual $R_s$, which highlights structural deviations from the
local context.
In Attention-guided Spatial Amplification, $R_s$ is passed through a two-layer
convolutional attention network $\omega_s = \sigma(\mathrm{Conv}(\mathrm{Conv}(R_s)))$
to produce a spatial attention weight, which is applied to modulate $V^\ell$ via
a residual connection, yielding $\mathbf{A}^\ell_{\text{spa}}$.

\vspace{-0.5ex}
\paragraph{Frequency branch.}
As shown in the right panel of Figure~\ref{fig:asa}, the frequency branch begins
with Multi-band Reweighting in the spectral domain.
$V^\ell$ is first transformed via 2D FFT into magnitude $\mathbf{E} =
|\mathrm{FFT}(V^\ell)|$ and phase $\boldsymbol{\Phi} = \angle\mathrm{FFT}(V^\ell)$.
The magnitude is separated into low- and high-frequency components via a circular
DC mask of radius $r$:
\begin{equation}
    \mathbf{E}_{\text{low}} = \mathbf{E} \odot \mathbf{M}_r, \quad
    \mathbf{E}_{\text{high}} = \mathbf{E} \odot (1 - \mathbf{M}_r),
    \label{eq:supp_mask}
\end{equation}
where $\mathbf{E}_{\text{low}}$ captures smooth background texture and
$\mathbf{E}_{\text{high}}$ captures fine-grained structural details.
Learnable weights $\omega_l$ and $\omega_h$ perform Spectrum Reweighting to produce
$\mathbf{E}' = \omega_l \cdot \mathbf{E}_{\text{low}} + \omega_h \cdot
\mathbf{E}_{\text{high}}$, which is reconstructed to the feature domain via 2D IFFT
with the original phase $\boldsymbol{\Phi}$ preserved.
In Spectrum-guided Frequency Amplification, the Textural Contrast Residual
$|V^\ell - \mathrm{IFFT}(\mathbf{E}' \cdot e^{j\boldsymbol{\Phi}})|$ highlights
regions with abnormal spectral patterns, and is concatenated with $V^\ell$ and
passed through a convolution to yield $\mathbf{A}^\ell_{\text{freq}}$.

\vspace{-0.5ex}
\paragraph{Adaptive fusion.}
The two branch outputs are combined via a learnable weight $\beta$:
\begin{equation}
    A^\ell = \beta \cdot \mathbf{A}^\ell_{\text{spa}} + (1-\beta) \cdot
    \mathbf{A}^\ell_{\text{freq}},
    \label{eq:supp_fusion}
\end{equation}
allowing the model to shift emphasis between structural and textural anomaly signals
depending on the defect category.
The amplified features $A^\ell$ are then passed to the VGTD module as enriched
visual contexts for drift generation.

\section{Breakdown Performance}
We provide the complete subset-wise performance breakdown in
Tables~\ref{tab:mvtec_1shot_auroc}--\ref{tab:mvtec_4shot_pro} for
MVTec-AD~\cite{bergmann2019mvtec} and
Tables~\ref{tab:visa_1shot_auroc}--\ref{tab:visa_4shot_pro} for
VisA~\cite{zou2022spot}, covering 1-shot, 2-shot, and 4-shot settings
across four metrics: AUROC, pAUROC, AUPR, and PRO.
In total, 24 tables are provided to facilitate detailed per-category
comparison against six baselines: PatchCore~\cite{roth2022towards},
WinCLIP~\cite{jeong2023winclip}, AnomalyGPT~\cite{gu2024anomalygpt},
PromptAD~\cite{li2024promptad}, KAG-prompt~\cite{tao2025kernel},
and FocusPatch-AD~\cite{ding2025focuspatch}.

On MVTec-AD, DriftAD consistently achieves the highest mean performance
across all shot settings and metrics.
Gains are particularly notable on structurally challenging categories
such as \textit{grid}, \textit{transistor}, and \textit{screw}, where
methods relying on static global text embeddings tend to produce diffuse
or misaligned anomaly maps.
For example, as shown in Table~\ref{tab:mvtec_1shot_auroc}, on \textit{grid} under the 1-shot setting, DriftAD achieves
$100.0\%$ AUROC compared to $99.1\%$ for KAG-prompt~\cite{tao2025kernel} and $99.1\%$ for
FocusPatch-AD~\cite{ding2025focuspatch}.
On \textit{screw}, DriftAD reaches $83.4\%$ AUROC, outperforming
KAG-prompt ($73.1\%$) and FocusPatch-AD ($78.4\%$) by substantial
margins, suggesting that VGTD is particularly
effective when defects are fine-grained and spatially localized.
The performance advantage is most pronounced under the 1-shot setting,
where limited support information makes the text-side adaptation of
VGTD especially beneficial.

On VisA, DriftAD surpasses all baselines across all metrics and shot
settings.
Gains are most consistent on fine-grained categories such as
\textit{macaroni1}, \textit{pcb1}, and \textit{pcb2}, where subtle
structural anomalies require precise spatial localization.
As shown in Table~\ref{tab:visa_1shot_auroc}, on \textit{pcb1} under 1-shot, DriftAD achieves $94.7\%$ AUROC, compared
to $86.2\%$ for FocusPatch-AD and $84.4\%$ for KAG-prompt.
On \textit{pcb2}, DriftAD reaches $87.4\%$ AUROC versus $70.9\%$ for
FocusPatch-AD, a gain of $16.5\%$, and versus $83.3\%$ for KAG-prompt.
These results confirm that layer-wise, spatially-adaptive anomaly descriptors provide substantially richer spatial guidance than patch-level feature
selection or static prompt learning.
The performance gap between DriftAD and competing methods narrows slightly
under the 4-shot setting, consistent with the general trend that additional
support images reduce the reliance on text-guided feature adaptation.

\section{More Qualitative Results}

\begin{figure*}[t]
    \centering
\includegraphics[width=0.89\textwidth]{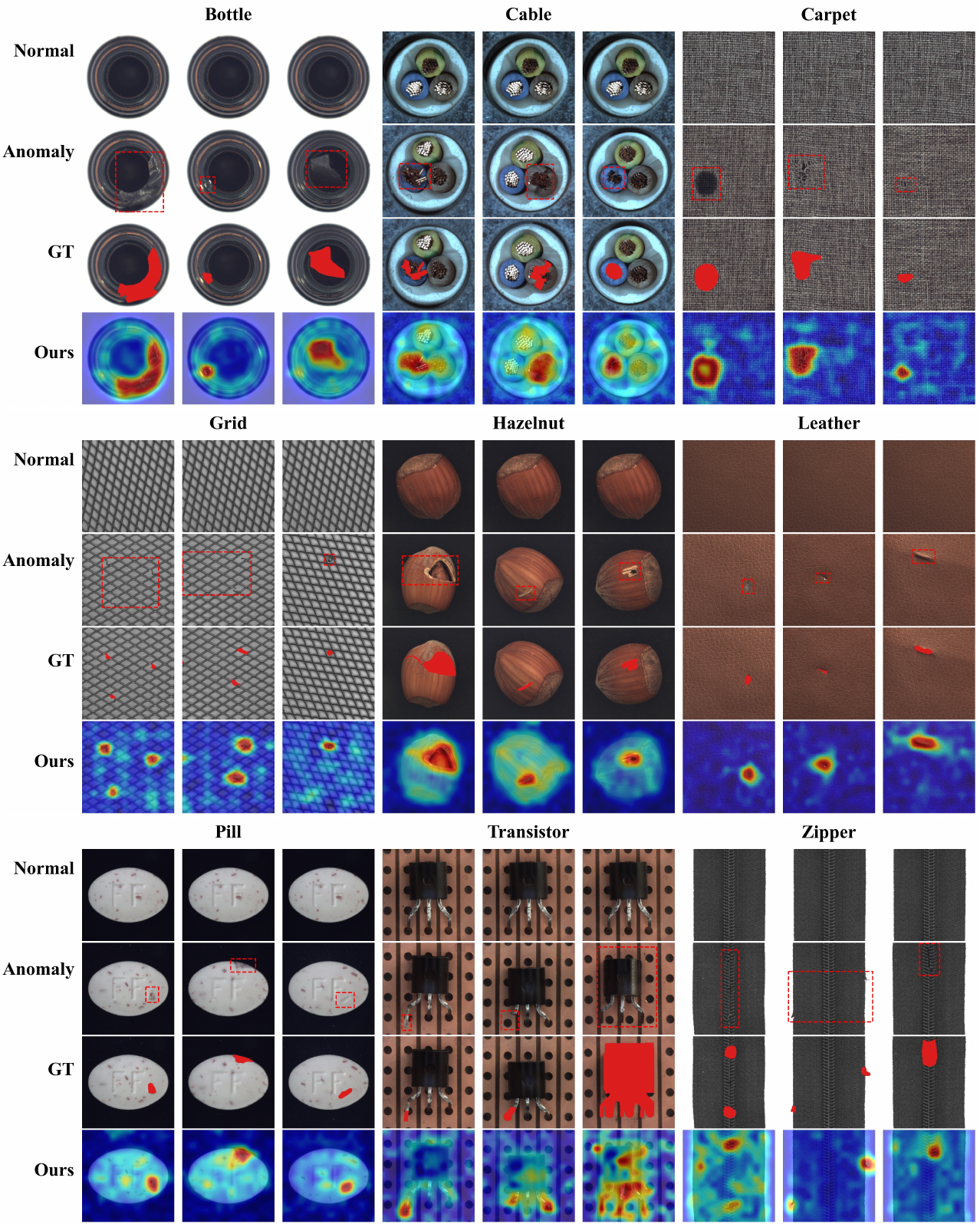}
    \caption{Qualitative anomaly localization results on selected categories of MVTec-AD. Each group presents three defect samples from the same category.} 
    \label{fig:vis_supply1}
\end{figure*}

\begin{figure*}[t]
    \centering
\includegraphics[width=0.89\textwidth]{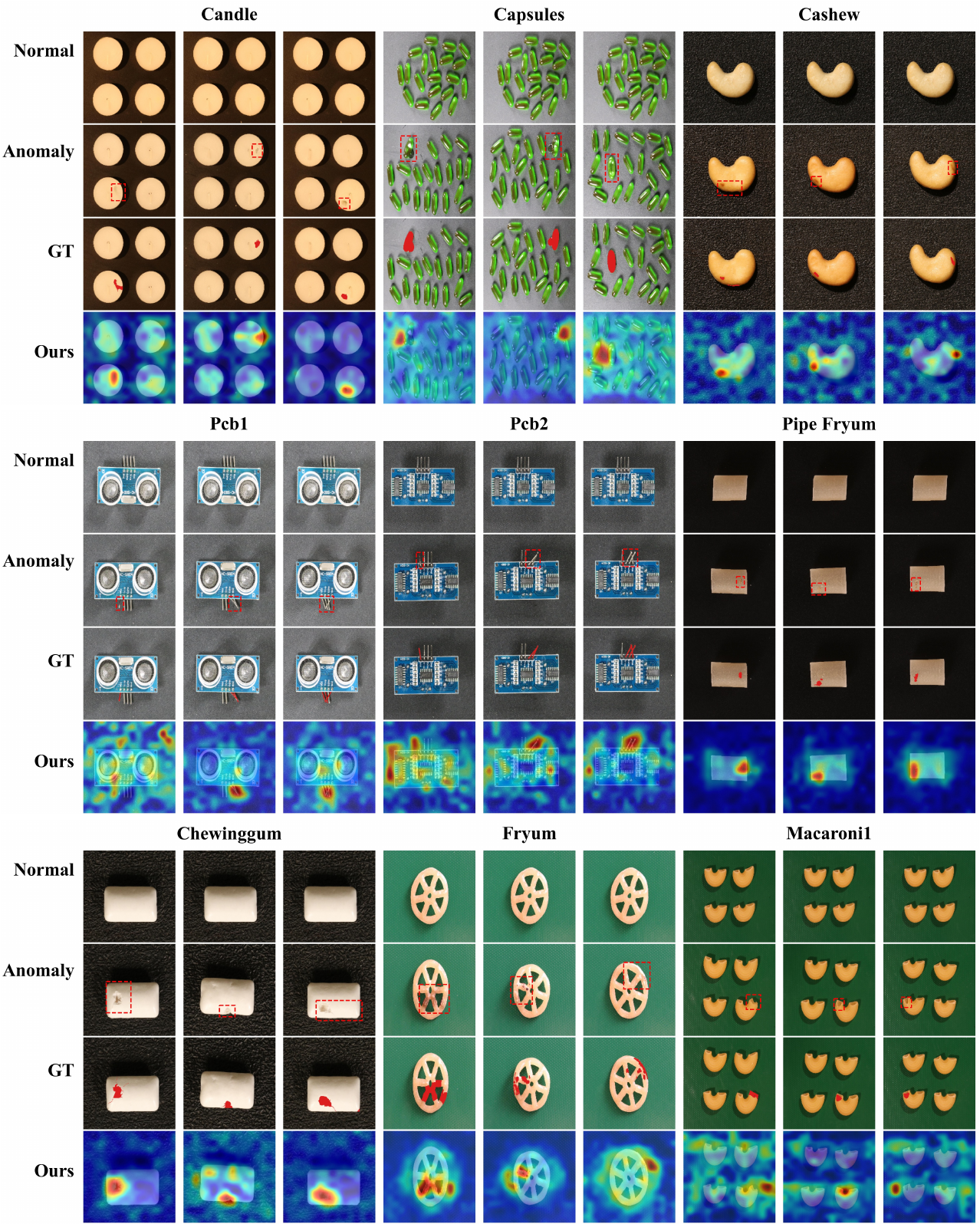}
    \caption{Qualitative anomaly localization results on selected categories of VisA. Each group presents three defect samples from the same category.} 
    \label{fig:vis_supply2}
\end{figure*}

We provide additional visualization of anomaly localization results on
MVTec-AD~\cite{bergmann2019mvtec} and VisA~\cite{zou2022spot} in
Figures~\ref{fig:vis_supply1}--\ref{fig:vis_supply2}.
For each dataset, we select representative categories and display three
samples per category, covering diverse defect types.
Each figure shows four rows: a normal reference image, an anomalous input
with the defect region highlighted by a red bounding box, the ground-truth
binary mask, and our predicted anomaly map.
The results demonstrate that DriftAD consistently produces sharp and
precise anomaly maps across a wide range of defect patterns, including
both structural and textural defects.

\begin{table*}[t]
\centering
\caption{Subset-wise performance comparison of the 1-shot setting for AUROC on MVTecAD.
}
\label{tab:mvtec_1shot_auroc}
\renewcommand{\arraystretch}{0.8}
\setlength{\tabcolsep}{6pt}
\begin{tabular}{lcccc ccc}
\toprule
MVTecAD & \multicolumn{7}{c}{1-shot} \\
\cmidrule(lr){2-8}
AUROC & PatchCore & WinCLIP & AnomalyGPT & PromptAD & KAG-prompt & FocusPatch-AD  & DriftAD (\textbf{Ours})\\
\midrule
bottle     & 99.4 & 98.2 & 98.1 & 99.8 & 98.7 & 99.8 & 100.0  \\
cable      & 88.8 & 88.9 & 88.8 & 94.2 & 90.7 & 92.6 & 93.7 \\
capsule    & 67.8 & 72.3 & 90.1 & 84.6 & 90.9 & 95.3 & 92.5\\
carpet     & 95.3 & 99.8 & 99.8 & 100.0 & 100.0 & 100.0 & 100.0\\
grid       & 63.6 & 99.5 & 98.4 & 99.8 & 99.1 & 99.1 & 100.0\\
hazelnut   & 88.3 & 97.5 & 100.0 & 99.8 & 100.0 & 98.8 & 99.8\\
leather    & 97.3 & 99.9 & 100.0 & 100.0 & 100.0 & 100.0 & 100.0\\
metal nut  & 73.4 & 98.7 & 98.3 & 99.1 & 99.9 &  97.8 & 100.0\\
pill       & 81.9 & 91.2 & 94.6 & 92.6 & 94.8 & 90.6 & 96.3\\
screw      & 44.4 & 86.4 & 74.7 & 65.0 & 73.1 & 78.4  & 83.4\\
tile       & 99.0 & 99.9 & 98.3 & 100.0 & 99.8 & 99.2 & 99.9\\
toothbrush & 83.3 & 92.2 & 98.4 & 98.9 & 99.2 & 99.7 & 99.4\\
transistor & 78.1 & 83.4 & 77.2 & 94.0 & 93.9 & 96.7 & 96.8\\
wood       & 97.8 & 99.9 & 98.1 & 97.9 & 98.8 & 99.6 & 98.5 \\
zipper     & 92.3 & 88.8 & 97.2 & 93.9 & 98.9 & 94.7 & 97.9\\
\midrule
mean       & 83.4 & 93.1 & 94.1 & 94.6 & 95.8 & \underline{96.0} & \textbf{97.2}\\
\bottomrule
\end{tabular}
\end{table*}

\begin{table*}[t]
\centering
\caption{Subset-wise performance comparison of the 2-shot setting for AUROC on MVTecAD.
}
\label{tab:mvtec_2shot_pauroc}
\renewcommand{\arraystretch}{0.8}
\setlength{\tabcolsep}{6pt}
\begin{tabular}{lcccc ccc}
\toprule
MVTecAD & \multicolumn{7}{c}{2-shot} \\
\cmidrule(lr){2-8}
AUROC & PatchCore & WinCLIP & AnomalyGPT & PromptAD & KAG-prompt & FocusPatch-AD & DriftAD (\textbf{Ours})\\
\midrule
bottle     & 99.2 & 99.3 & 99.4 & 100.0 & 99.6 & 99.9 & 99.9 \\
cable      & 91.0 & 88.4 & 90.4 & 99.9  & 91.2 & 93.0 & 94.0 \\
capsule    & 72.8 & 77.3 & 89.8 & 100.0 & 91.1 & 90.4 & 95.6 \\
carpet     & 96.6 & 99.8 & 100.0 & 100.0 & 100.0 & 100.0 & 100.0\\
grid       & 67.7 & 99.4 & 99.4 & 98.7  & 99.2 & 99.1 & 100.0\\
hazelnut   & 93.2 & 98.3 & 100.0 & 99.4  & 100.0 & 98.9 & 99.8\\
leather    & 97.9 & 99.9 & 100.0 & 91.8  & 100.0 & 100.0 & 100.0\\
metal nut  & 77.7 & 99.4 & 100.0 & 91.3  & 99.9 & 99.9 & 100.0 \\
pill       & 82.9 & 92.3 & 95.9  & 100.0 & 94.6 & 92.9 & 96.3\\
screw      & 49.0 & 86.0 & 83.5  & 98.6  & 82.4 & 70.1 & 85.4\\
tile       & 98.5 & 99.9 & 99.5  & 93.6  & 99.9 & 100.0 & 99.9\\
toothbrush & 85.9 & 97.5 & 99.2  & 71.0  & 98.9 & 97.2 & 99.2\\
transistor & 90.0 & 85.3 & 80.4  & 97.5  & 95.0 & 95.9 & 97.5\\
wood       & 98.3 & 99.9 & 99.1  & 97.4  & 98.4 & 99.1 & 98.5\\
zipper     & 94.0 & 94.0 & 96.2  & 95.8  & 98.5 & 96.2 & 99.6\\
\midrule
mean        & 86.3 & 94.4 & 95.5 & 95.7 & \underline{96.6} & 96.4 & \textbf{97.7}\\
\bottomrule
\end{tabular}
\end{table*}
\begin{table*}[t]
\centering
\caption{Subset-wise performance comparison of the 4-shot setting for AUROC on MVTecAD.
}
\label{tab:mvtec_4shot_pauroc}
\setlength{\tabcolsep}{6pt}
\renewcommand{\arraystretch}{0.8}
\begin{tabular}{lcccc ccc}
\toprule
MVTecAD & \multicolumn{5}{c}{4-shot} \\
\cmidrule(lr){2-8}
AUROC & PatchCore & WinCLIP & AnomalyGPT & PromptAD & KAG-prompt & FocusPatch-AD & DriftAD (\textbf{Ours}) \\
\midrule
bottle     & 99.2 & 99.3 & 99.4 & 100.0 & 99.5 & 100.0 & 100.0  \\
cable      & 91.0 & 90.9 & 91.4 & 98.8  & 91.5 & 94.4 & 94.5 \\
capsule    & 72.8 & 82.3 & 91.1 & 100.0 & 92.9 & 91.7 & 96.1\\
carpet     & 96.6 & 100.0 & 100.0 & 100.0 & 100.0 & 100.0 & 100.0 \\
grid       & 67.7 & 99.6 & 99.4 & 98.6  & 99.2 & 98.4 & 100.0 \\
hazelnut   & 93.2 & 98.4 & 100.0 & 99.8  & 100.0 & 99.6 & 	100.0\\
leather    & 97.9 & 100.0 & 100.0 & 95.4  & 100.0 & 100.0 & 100.0 \\
metal nut  & 77.7 & 99.5 & 100.0 & 91.5  & 100.0 & 99.8 & 100.0 \\
pill       & 82.9 & 92.8 & 95.8  & 99.8  & 95.1 & 92.6 & 96.3\\
screw      & 49.0 & 87.9 & 84.8  & 100.0 & 84.7 & 81.0 & 88.4 \\
tile       & 98.5 & 99.9 & 99.2  & 92.9  & 99.9 & 99.9 & 99.9 \\
toothbrush & 85.9 & 96.7 & 98.1  & 83.6  & 99.2 & 98.3 & 99.2 \\
transistor & 90.0 & 85.7 & 88.3  & 98.1  & 96.9 & 96.6 & 97.9\\
wood       & 98.3 & 99.8 & 99.5  & 95.6  & 98.6 & 99.3 & 98.4\\
zipper     & 94.0 & 94.5 & 97.5  & 95.0  & 99.1 & 95.2 & 99.9\\
\midrule
mean       & 86.3 & 95.2 & 96.3 & 96.6 & \underline{97.1} &  97.4 & 	\textbf{98.0} \\
\bottomrule
\end{tabular}
\end{table*}
\begin{table*}[t]
\centering
\caption{Subset-wise performance comparison of the 1-shot setting for pAUROC on MVTecAD.}
\label{tab:mvtec_1shot_pauroc}
\setlength{\tabcolsep}{6pt}
\renewcommand{\arraystretch}{0.8}
\begin{tabular}{lcccc ccc}
\toprule
MVTecAD & \multicolumn{7}{c}{1-shot} \\
\cmidrule(lr){2-8}
pAUROC & PatchCore & WinCLIP & AnomalyGPT & PromptAD & KAG-prompt & FocusPatch-AD & DriftAD (\textbf{Ours}) \\
\midrule
bottle     & 97.9 & 97.5 & 97.6 & 99.6 & 97.5 & 97.4 & 98.3\\
cable      & 95.5 & 93.8 & 93.1 & 98.4 & 91.6 & 81.6 & 95.1\\
capsule    & 95.6 & 94.6 & 92.3 & 99.5 & 97.4 & 95.0 & 97.1\\
carpet     & 98.4 & 99.4 & 99.1 & 95.9 & 99.4 & 99.7 & 99.6\\
grid       & 58.8 & 96.8 & 95.9 & 95.1 & 97.0 & 97.6 & 98.3\\
hazelnut   & 95.8 & 98.5 & 98.6 & 97.3 & 98.5 & 99.6 & 99.0\\
leather    & 98.8 & 99.3 & 99.3 & 93.3 & 99.5 & 99.8 & 99.6\\
metal nut  & 89.3 & 90.0 & 90.6 & 97.3 & 93.7 & 88.7 & 92.9\\
pill       & 93.1 & 96.4 & 95.9 & 98.4 & 96.8 & 93.1 & 97.1\\
screw      & 89.6 & 94.5 & 94.8 & 91.4 & 96.5 & 92.8 & 98.5\\
tile       & 94.1 & 96.3 & 96.0 & 92.8 & 97.7 & 99.0 & 98.0\\
toothbrush & 97.3 & 97.8 & 98.2 & 94.0 & 98.6 & 98.6 & 98.9\\
transistor & 84.9 & 85.0 & 87.6 & 99.1 & 84.6 & 88.8 & 85.7\\
wood       & 92.7 & 94.6 & 96.4 & 89.4 & 96.4 & 94.7 & 96.9\\
zipper     & 97.4 & 93.9 & 93.8 & 96.6 & 97.3 & 91.7 & 96.6\\
\midrule
mean       & 92.0 & 95.2 & 95.3 & 95.9 &  96.2 & \underline{96.4} & \textbf{96.8}\\
\bottomrule
\end{tabular}
\end{table*}
\begin{table*}[t]
\centering
\caption{Subset-wise performance comparison of the 2-shot setting for pAUROC on MVTecAD.
}
\label{tab:mvtec_2shot_pauroc}
\setlength{\tabcolsep}{6pt}
\renewcommand{\arraystretch}{0.8}
\begin{tabular}{lcccc ccc}
\toprule
MVTecAD & \multicolumn{7}{c}{2-shot} \\
\cmidrule(lr){2-8}
pAUROC & PatchCore & WinCLIP & AnomalyGPT & PromptAD & KAG-prompt & FocusPatch-AD & DriftAD (\textbf{Ours})\\
\midrule
bottle     & 98.1 & 97.7 & 97.7 & 99.5 & 97.7 & 97.9 & 98.5 \\
cable      & 96.4 & 94.3 & 93.2 & 97.6 & 92.4 & 82.8 & 95.3\\
capsule    & 96.5 & 96.4 & 92.3 & 99.3 & 97.7 & 93.1 & 97.5\\
carpet     & 98.5 & 99.3 & 99.4 & 96.1 & 99.5 & 99.8 & 99.6\\
grid       & 62.6 & 97.7 & 96.5 & 95.5 & 97.1 & 96.9 & 98.4\\
hazelnut   & 96.3 & 98.7 & 98.7 & 97.6 & 98.5 & 99.6 & 99.0\\
leather    & 99.0 & 99.3 & 99.3 & 93.2 & 99.5 & 99.9 & 99.6\\
metal nut  & 94.6 & 91.4 & 91.6 & 97.4 & 94.5 & 91.4 & 93.1 \\
pill       & 94.2 & 97.0 & 96.9 & 98.5 & 96.9 & 93.4 & 97.2\\
screw      & 90.0 & 95.2 & 95.6 & 95.1 & 97.3 & 95.0 & 98.9\\
tile       & 94.4 & 96.5 & 96.2 & 94.1 & 97.8 & 99.1 & 98.0\\
toothbrush & 97.5 & 98.1 & 98.3 & 95.5 & 98.7 & 98.8 & 98.9\\
transistor & 89.6 & 88.3 & 88.4 & 99.0 & 86.1 & 90.7 & 87.7\\
wood       & 93.2 & 95.3 & 96.4 & 89.1 & 96.3 & 94.5 & 97.0\\
zipper     & 98.0 & 94.1 & 93.9 & 95.5 & 97.3 & 91.8 & 97.0\\
\midrule
mean       & 93.3 & 96.0 & 95.6 & 96.2 & 96.5 & \textbf{96.9} & \textbf{97.0}\\
\bottomrule
\end{tabular}
\end{table*}
\begin{table*}[t]
\centering
\caption{Subset-wise performance comparison of the 4-shot setting for pAUROC on MVTecAD.}
\label{tab:mvtec_4shot_pauroc}
\setlength{\tabcolsep}{6pt}
\renewcommand{\arraystretch}{0.8}
\begin{tabular}{lcccc ccc}
\toprule
MVTecAD & \multicolumn{7}{c}{4-shot} \\
\cmidrule(lr){2-8}
pAUROC & PatchCore & WinCLIP & AnomalyGPT & PromptAD & KAG-prompt & FocusPatch-AD & DriftAD (\textbf{Ours}) \\
\midrule
bottle     & 98.2 & 97.8 & 98.2 & 99.5 & 98.0 & 97.9 & 98.6 \\
cable      & 97.5 & 94.9 & 94.3 & 98.2 & 92.5 & 83.0 & 95.4\\
capsule    & 96.8 & 96.2 & 93.4 & 99.3 & 97.9 & 95.4 & 97.5 \\
carpet     & 98.6 & 99.3 & 99.4 & 96.2 & 99.5 & 99.8 & 99.6\\
grid       & 69.4 & 98.0 & 97.6 & 95.2 & 97.6 & 97.8 & 98.4\\
hazelnut   & 97.6 & 98.8 & 98.8 & 97.9 & 98.7 & 99.7 & 99.0\\
leather    & 99.1 & 99.3 & 99.3 & 93.9 & 99.5 & 99.9 & 99.6\\
metal nut  & 95.9 & 92.9 & 93.4 & 97.7 & 94.6 & 92.4 & 94.3\\
pill       & 94.8 & 97.1 & 97.2 & 98.5 & 97.0 & 94.1 & 97.3\\
screw      & 91.3 & 96.0 & 96.8 & 94.9 & 97.5 & 95.2 & 99.1\\
tile       & 94.6 & 96.6 & 96.3 & 94.1 & 97.8 & 99.9 & 98.0\\
toothbrush & 98.4 & 98.4 & 98.3 & 96.2 & 98.7 & 98.9 & 98.9 \\
transistor & 90.7 & 88.5 & 90.2 & 99.0 & 86.6 & 89.0 & 88.0\\
wood       & 93.5 & 95.4 & 96.4 & 90.6 & 96.2 & 94.7 & 97.0\\
zipper     & 98.1 & 94.2 & 94.1 & 96.6 & 97.7 & 92.1 & 97.1\\
\midrule
mean       & 94.3 & 96.2 & 96.2 & 96.5 & \underline{96.7} & \textbf{97.2} & \textbf{97.2} \\
\bottomrule
\end{tabular}
\end{table*}
\begin{table*}[t]
\centering
\caption{Subset-wise performance comparison of the 1-shot setting for AUPR on MVTecAD.}
\label{tab:mvtec_1shot_aupr}
\setlength{\tabcolsep}{6pt}
\renewcommand{\arraystretch}{0.8}
\begin{tabular}{lcccc ccc}
\toprule
MVTecAD & \multicolumn{7}{c}{1-shot} \\
\cmidrule(lr){2-8}
AUPR & PatchCore & WinCLIP & AnomalyGPT & PromptAD & KAG-prompt & FocusPatch-AD & DriftAD (Ours)\\
\midrule
bottle     & 99.8 & 99.4 & 98.7 & 99.8  & 99.6 & 100.0 & 100.0 \\
cable      & 93.8 & 93.2 & 93.1 & 95.5  & 94.5 & 96.0 & 96.9 \\
capsule    & 89.4 & 91.6 & 97.8 & 97.8  & 98.0 & 97.7 & 98.5 \\
carpet     & 98.7 & 99.9 & 99.9 & 100.0 & 100.0 & 100.0 & 100.0\\
grid       & 81.1 & 99.9 & 98.6 & 98.8  & 99.7 & 99.7 & 100.0\\
hazelnut   & 92.9 & 98.6 & 100.0 & 99.7 & 100.0 & 99.3 & 99.9\\
leather    & 99.1 & 100.0 & 100.0 & 100.0 & 100.0 & 100.0 & 100.0 \\
metal nut  & 91.0 & 99.7 & 98.4 & 99.6  & 100.0 & 99.5 & 100.0\\
pill       & 96.5 & 98.3 & 98.7 & 98.5  & 99.1 & 98.1 & 99.3\\
screw      & 71.4 & 94.2 & 88.2 & 78.5  & 89.0 & 83.7 & 94.0\\
tile       & 99.6 & 100.0 & 99.4 & 100.0 & 99.9 & 99.7 & 100.0\\
toothbrush & 93.5 & 96.7 & 99.1 & 98.5  & 99.7 & 99.9 & 99.8\\
transistor & 77.7 & 79.0 & 68.5 & 91.2  & 92.1 & 88.2 & 95.5\\
wood       & 99.3 & 100.0 & 98.5 & 99.6  & 99.6 & 99.9 & 99.5\\
zipper     & 97.2 & 96.8 & 99.0 & 99.0  & 99.7 & 98.2 & 99.4\\
\midrule
mean       & 92.2 & 96.5 & 95.9 & 97.1 & \underline{98.1} & 97.3 & \textbf{98.8} \\
\bottomrule
\end{tabular}
\end{table*}
\begin{table*}[t]
\centering
\caption{Subset-wise performance comparison of the 2-shot setting for AUPR on MVTecAD.}
\label{tab:mvtec_2shot_aupr}
\setlength{\tabcolsep}{6pt}
\renewcommand{\arraystretch}{0.8}
\begin{tabular}{lcccc ccc}
\toprule
MVTecAD & \multicolumn{7}{c}{2-shot} \\
\cmidrule(lr){2-8}
AUPR & PatchCore & WinCLIP & AnomalyGPT & PromptAD & KAG-prompt & FocusPatch-AD & DriftAD (\textbf{Ours}) \\
\midrule
bottle     & 99.8 & 99.8 & 99.8  & 99.9  & 99.9 & 100.0 & 100.0\\
cable      & 95.1 & 92.9 & 93.9  & 96.9  & 95.0 & 96.5 & 97.0\\
capsule    & 91.0 & 93.3 & 97.7  & 97.0  & 98.0 & 97.8 & 99.1\\
carpet     & 99.0 & 99.9 & 100.0 & 100.0 & 100.0 & 99.9 & 100.0\\
grid       & 84.1 & 99.8 & 99.8  & 99.9  & 99.7 & 99.7 & 100.0 \\
hazelnut   & 96.0 & 99.1 & 100.0 & 99.8  & 100.0 & 98.7& 99.90\\
leather    & 99.3 & 100.0 & 100.0 & 100.0 & 100.0 & 99.9 &100.0\\
metal nut  & 92.3 & 99.9 & 100.0 & 100.0 & 100.0 & 100.0 & 100.0 \\
pill       & 96.6 & 98.6 & 99.3  & 97.8  & 99.0 & 98.7 & 97.0\\
screw      & 72.9 & 94.1 & 94.3  & 86.7  & 94.0 & 84.3 & 94.6\\
tile       & 99.4 & 100.0 & 99.8 & 100.0 & 99.9 & 100.0 & 100.0 \\
toothbrush & 94.1 & 99.0 & 99.7  & 99.3  & 99.6 & 99.0 & 99.7\\
transistor & 89.3 & 80.7 & 69.3  & 92.2  & 93.6 & 94.1 & 96.5\\
wood       & 99.5 & 100.0 & 99.7 & 99.7  & 99.5 & 99.8 & 99.5 \\
zipper     & 97.8 & 98.3 & 99.0  & 99.3  & 99.6 & 99.0 & 99.90 \\
\midrule
mean       & 93.8 & 97.0 & 96.8 & 97.9 & \underline{98.5} & 97.8 & \textbf{99.0}\\
\bottomrule
\end{tabular}
\end{table*}
\begin{table*}[t]
\centering
\caption{Subset-wise performance comparison of the 4-shot setting for AUPR on MVTecAD.}
\label{tab:mvtec_4shot_aupr}
\setlength{\tabcolsep}{6pt}
\renewcommand{\arraystretch}{0.8}
\begin{tabular}{lcccc ccc}
\toprule
MVTecAD & \multicolumn{7}{c}{4-shot} \\
\cmidrule(lr){2-8}
AUPR & PatchCore & WinCLIP & AnomalyGPT & PromptAD & KAG-prompt & FocusPatch-AD & DriftAD (\textbf{Ours}) \\
\midrule
bottle     & 99.8 & 99.8 & 99.8  & 100.0 & 99.9 & 100.0 & 100.0\\
cable      & 97.1 & 94.4 & 95.4  & 97.4  & 95.3 & 97.0 &97.2 \\
capsule    & 94.9 & 95.1 & 98.5  & 98.6  & 98.4 & 98.2 & 99.2\\
carpet     & 98.8 & 100.0 & 100.0 & 100.0 & 100.0 & 100.0 & 100.0\\
grid       & 86.4 & 99.9 & 99.7  & 99.7  & 99.7  & 99.5 & 100.0\\
hazelnut   & 97.0 & 99.1 & 100.0 & 99.9  & 100.0 & 99.8 & 99.9\\
leather    & 99.6 & 100.0 & 100.0 & 100.0 & 100.0 & 100.0 & 100.0\\
metal nut  & 97.0 & 99.9 & 100.0 & 99.9  & 100.0 & 100.0 & 100.0 \\
pill       & 96.9 & 98.6 & 99.3  & 98.5  & 99.1 & 98.6 & 99.3\\
screw      & 71.8 & 94.9 & 94.5  & 93.8  & 94.7 & 93.2 & 94.6 \\
tile       & 99.6 & 100.0 & 99.7 & 100.0 & 99.9 & 100.0 & 100.0 \\
toothbrush & 94.8 & 98.7 & 98.3  & 99.7  & 99.7 & 99.4 & 99.7\\
transistor & 84.5 & 80.7 & 83.1  & 92.2  & 96.5 & 94.8 & 97.3\\
wood       & 99.5 & 99.9 & 99.0  & 99.5  & 99.5 & 99.8 & 99.5 \\
zipper     & 99.5 & 98.5 & 97.3  & 98.5  & 99.8 & 98.5 & 99.9 \\
\midrule
mean       & 94.5 & 97.3 & 97.6 & 98.5 & \underline{98.8} & 98.6 &  \textbf{99.2} \\
\bottomrule
\end{tabular}
\end{table*}

\begin{table*}[t]
\centering
\caption{Subset-wise performance comparison of the 1-shot setting for PRO on MVTecAD.}
\label{tab:mvtec_1shot_pro}
\setlength{\tabcolsep}{6pt}
\renewcommand{\arraystretch}{0.8}
\begin{tabular}{lcccc ccc}
\toprule
MVTecAD & \multicolumn{7}{c}{1-shot} \\
\cmidrule(lr){2-8}
PRO & PatchCore & WinCLIP & AnomalyGPT & PromptAD & KAG-prompt & FocusPatch-AD & DriftAD (\textbf{Ours}) \\
\midrule
bottle     & 93.5 & 91.2 & 94.4 & 93.6 & 93.9 & 94.8 & 94.1 \\
cable      & 84.7 & 72.5 & 83.5 & 87.3 & 80.8 & 83.0 & 85.8\\
capsule    & 83.9 & 85.6 & 85.5 & 80.1 & 93.6 & 88.3 & 94.3\\
carpet     & 93.3 & 97.4 & 96.8 & 98.3 & 97.3 & 97.8 & 98.1 \\
grid       & 21.7 & 90.5 & 90.9 & 94.3 & 92.1 & 93.0 & 94.4\\
hazelnut   & 88.3 & 93.7 & 95.6 & 92.9 & 94.2 & 91.1 & 96.6\\
leather    & 95.2 & 98.6 & 97.9 & 98.7 & 98.1 & 98.3 & 98.7\\
metal nut  & 66.7 & 84.7 & 87.0 & 83.1 & 91.5 & 89.5 & 92.7\\
pill       & 89.5 & 93.5 & 94.6 & 90.8 & 96.2 & 90.6 & 96.8\\
screw      & 68.1 & 82.3 & 85.9 & 78.1 & 88.4 & 83.9 & 93.0\\
tile       & 82.5 & 89.4 & 90.6 & 90.7 & 93.5 & 91.4 & 94.0\\
toothbrush & 79.0 & 85.3 & 90.8 & 90.1 & 91.7 & 90.2 & 88.3\\
transistor & 70.9 & 65.0 & 71.2 & 67.5 & 65.0 & 71.2 & 70.2 \\
wood       & 87.1 & 91.0 & 93.2 & 92.4 & 93.4 & 90.8 & 94.2\\
zipper     & 91.2 & 86.0 & 84.8 & 81.0 & 91.7 & 80.0 & 91.4\\
\midrule
mean       & 79.3 & 87.1 & 89.5 & 87.9 & \underline{90.8} & 88.9 & \textbf{92.2} \\
\bottomrule
\end{tabular}
\end{table*}
\begin{table*}[t]
\centering
\caption{Subset-wise performance comparison of the 2-shot setting for PRO on MVTecAD.}
\label{tab:mvtec_2shot_pro}
\setlength{\tabcolsep}{6pt}
\renewcommand{\arraystretch}{0.8}
\begin{tabular}{lcccc ccc}
\toprule
MVTecAD & \multicolumn{7}{c}{2-shot} \\
\cmidrule(lr){2-8}
PRO & PatchCore & WinCLIP & AnomalyGPT & PromptAD & KAG-prompt & FocusPatch-AD & DriftAD (\textbf{Ours})\\
\midrule
bottle     & 93.9 & 91.8 & 94.7 & 93.9 & 94.4 & 94.0 & 94.7\\
cable      & 88.5 & 74.7 & 84.2 & 87.8 & 81.7 & 82.3 & 86.5 \\
capsule    & 86.6 & 90.6 & 85.7 & 79.2 & 94.3 & 89.1 & 95.1\\
carpet     & 93.7 & 97.3 & 97.3 & 98.2 & 97.5 & 97.9 & 98.2\\
grid       & 23.7 & 92.8 & 91.3 & 95.0 & 92.0 & 93.6 & 94.4\\
hazelnut   & 89.8 & 94.2 & 95.7 & 93.4 & 94.2 & 93.8 & 96.5\\
leather    & 95.9 & 98.3 & 97.9 & 98.7 & 98.0 & 98.6 & 98.6\\
metal nut  & 79.6 & 86.7 & 87.8 & 87.7 & 92.3 & 89.0 & 92.9\\
pill       & 91.6 & 94.5 & 95.6 & 90.5 & 96.3 & 90.6 & 97.0\\
screw      & 69.0 & 84.1 & 86.0 & 74.7 & 88.9 & 83.1 & 94.6\\
tile       & 82.5 & 89.6 & 90.7 & 90.9 & 93.6 & 91.0 & 94.0\\
toothbrush & 81.0 & 84.7 & 90.9 & 91.6 & 91.9 & 89.3 & 88.4\\
transistor & 78.8 & 68.6 & 73.5 & 68.1 & 67.1 & 72.1 & 72.2\\
wood       & 86.8 & 91.8 & 93.4 & 91.6 & 92.8 & 91.6 & 94.4 \\
zipper     & 92.8 & 86.4 & 84.9 & 86.4 & 91.5 & 80.5 & 92.0\\
\midrule
mean       & 82.3 & 88.4 & 90.0 & 88.5 & \underline{91.1} & 89.1 & \textbf{92.6}\\
\bottomrule
\end{tabular}
\end{table*}
\begin{table*}[t]
\centering
\caption{Subset-wise performance comparison of the 4-shot setting for PRO on MVTecAD.}
\label{tab:mvtec_4shot_pro}
\setlength{\tabcolsep}{6pt}
\renewcommand{\arraystretch}{0.8}
\begin{tabular}{lcccc ccc}
\toprule
MVTecAD & \multicolumn{7}{c}{4-shot} \\
\cmidrule(lr){2-8}
PRO & PatchCore & WinCLIP & AnomalyGPT & PromptAD & KAG-prompt & FocusPatch-AD & DriftAD (\textbf{Ours}) \\
\midrule
bottle     & 94.0 & 91.6 & 95.2 & 94.5 & 94.6 & 93.9 & 94.9\\
cable      & 91.7 & 77.0 & 85.6 & 88.9 & 82.0 & 83.8 & 86.3\\
capsule    & 87.8 & 90.1 & 87.1 & 88.7 & 94.9 & 89.2 & 95.2 \\
carpet     & 93.9 & 97.0 & 97.3 & 98.2 & 97.4 & 98.0 & 98.0\\
grid       & 30.4 & 93.6 & 93.5 & 93.8 & 93.3 & 93.9 & 94.4\\
hazelnut   & 92.0 & 94.2 & 96.3 & 95.2 & 94.7 & 93.4 & 96.9\\
leather    & 96.4 & 98.0 & 97.7 & 98.4 & 97.9 & 96.7 & 98.7 \\
metal nut  & 83.8 & 89.4 & 90.1 & 87.6 & 92.4 & 88.5 & 93.4\\
pill       & 92.5 & 94.6 & 96.1 & 92.0 & 96.5 & 91.4 & 96.9 \\
screw      & 72.4 & 86.3 & 87.6 & 86.7 & 89.8 & 83.8 & 95.2\\
tile       & 83.0 & 89.9 & 90.4 & 90.9 & 93.2 & 91.2 & 93.8 \\
toothbrush & 85.5 & 86.0 & 90.9 & 91.3 & 91.9 & 90.4 & 88.8\\
transistor & 79.5 & 69.0 & 74.9 & 73.0 & 67.4 & 70.7 & 71.5 \\
wood       & 87.7 & 91.7 & 92.6 & 91.4 & 92.7 & 91.8 & 94.3\\
zipper     & 93.4 & 86.9 & 85.3 & 87.5 & 92.7 & 82.3 & 92.6\\
\midrule
mean       & 84.3 & 89.0 & 90.7 & 90.5 & \underline{91.4} & 89.3 & \textbf{92.7} \\
\bottomrule
\end{tabular}
\end{table*}

\begin{table*}[t]
\centering
\caption{Subset-wise performance comparison of the 1-shot setting for AUROC on VisA.}
\label{tab:visa_1shot_auroc}
\setlength{\tabcolsep}{6pt}
\renewcommand{\arraystretch}{0.85}
\begin{tabular}{lcccc ccc}
\toprule
VisA & \multicolumn{7}{c}{1-shot} \\
\cmidrule(lr){2-8}
AUROC & PatchCore & WinCLIP & AnomalyGPT & PromptAD & KAG-prompt & FocusPatch-AD & DriftAD (\textbf{Ours}) \\
\midrule
candle      & 85.1 & 93.4 & 85.8 & 90.3 & 96.3 & 92.7 & 94.0\\
capsules    & 60.0 & 85.0 & 85.8  & 84.5 & 89.1 & 82.6 & 90.6\\
cashew      & 89.5 & 94.0 & 91.5  & 95.6 & 94.4 & 92.2 & 96.7\\
chewinggum  & 97.3 & 97.6 & 98.0  & 96.4 & 98.7 & 96.8 & 99.6\\
fryum       & 75.0 & 88.5 & 92.2  & 90.3 & 93.7 & 82.5 & 92.5\\
macaroni1   & 68.0 & 82.9 & 89.9  & 88.6 & 96.6 & 86.4 & 96.3\\
macaroni2   & 55.6 & 70.2 & 84.7  & 69.1 & 84.1 & 69.8 & 86.1\\
pcb1        & 78.9 & 75.6 & 82.6  & 88.7 & 84.4 & 86.2 & 94.7\\
pcb2        & 81.5 & 62.2 & 75.6  & 71.6 & 83.3 & 70.9 & 87.4\\
pcb3        & 82.7 & 74.1 & 75.0  & 97.1 & 81.4 & 80.4 & 81.5\\
pcb4        & 93.9 & 85.2 & 88.8  & 91.4 & 97.9 & 92.1 & 98.6 \\
pipe fryum  & 90.7 & 97.2 & 99.3  & 96.9 & 99.5 & 96.3 & 99.1\\
\midrule
mean        & 79.9 & 83.8 & 87.4 & 86.9 & \underline{91.6} & 91.0 & \textbf{93.1} \\
\bottomrule
\end{tabular}
\end{table*}

\begin{table*}[t]
\centering
\caption{Subset-wise performance comparison of the 2-shot setting for AUROC on VisA.}
\label{tab:visa_2shot_auroc}
\setlength{\tabcolsep}{6pt}
\renewcommand{\arraystretch}{0.85}
\begin{tabular}{lcccc ccc}
\toprule
VisA & \multicolumn{7}{c}{2-shot} \\
\cmidrule(lr){2-8}
AUROC & PatchCore & WinCLIP & AnomalyGPT & PromptAD & KAG-prompt & FocusPatch-AD & DriftAD (\textbf{Ours}) \\
\midrule
candle      & 85.3 & 94.8 & 83.1 & 91.0 & 94.5 & 95.0 & 93.2 \\
capsules    & 57.8 & 84.9 & 88.8 & 84.9 & 89.9 & 81.4 & 92.7 \\
cashew      & 93.6 & 94.3 & 93.2 & 94.7 & 94.3 & 89.4 & 95.1 \\
chewinggum  & 97.8 & 97.3 & 98.4 & 96.6 & 98.8 & 96.0 & 99.2 \\
fryum       & 83.4 & 90.5 & 94.1 & 89.2 & 95.1 & 87.2 & 95.3 \\
macaroni1   & 75.6 & 83.3 & 91.4 & 84.2 & 95.9 & 84.4 & 94.8 \\
macaroni2   & 57.3 & 71.8 & 85.8 & 82.6 & 86.9 & 74.3 & 86.2 \\
pcb1        & 71.5 & 76.7 & 83.6 & 90.9 & 88.8 & 90.0 & 92.3 \\
pcb2        & 84.3 & 62.6 & 78.3 & 73.0 & 82.0 & 74.6 & 89.6 \\
pcb3        & 84.8 & 78.8 & 79.2 & 76.2 & 89.0 & 78.8 & 83.1 \\
pcb4        & 94.3 & 82.3 & 88.4 & 97.5 & 97.2 & 96.6 & 98.9 \\
pipe fryum  & 93.5 & 98.0 & 99.4 & 98.9 & 99.7 & 99.1 & 99.2 \\
\midrule
mean        & 81.6 & 84.6 & 88.6 & 88.3 & 92.7 & \underline{92.8} & \textbf{93.3} \\
\bottomrule
\end{tabular}
\end{table*}
\begin{table*}[t]
\centering
\caption{Subset-wise performance comparison of the 4-shot setting for AUROC on VisA.}
\label{tab:visa_4shot_auroc}
\setlength{\tabcolsep}{6pt}
\renewcommand{\arraystretch}{0.85}
\begin{tabular}{lccccc cc}
\toprule
VisA & \multicolumn{7}{c}{4-shot} \\
\cmidrule(lr){2-8}
AUROC & PatchCore & WinCLIP & AnomalyGPT & PromptAD & KAG-prompt & FocusPatch-AD & DriftAD (\textbf{Ours}) \\
\midrule
candle      & 87.8 & 95.1 & 85.6 & 93.0 & 95.0 & 90.2 & 94.3 \\
capsules    & 63.4 & 86.8 & 89.0 & 80.6 & 90.1 & 87.9 & 91.5 \\
cashew      & 93.0 & 95.2 & 94.2 & 93.6 & 97.3 & 95.3 & 96.2 \\
chewinggum  & 98.3 & 97.7 & 98.7 & 96.8 & 98.4 & 97.2 & 99.2 \\
fryum       & 88.6 & 90.8 & 95.2 & 89.0 & 94.3 & 88.9 & 94.7 \\
macaroni1   & 82.9 & 85.2 & 92.1 & 88.2 & 97.6 & 83.8 & 97.4 \\
macaroni2   & 61.7 & 70.9 & 87.2 & 81.2 & 87.2 & 77.2 & 84.5 \\
pcb1        & 84.7 & 88.3 & 83.1 & 90.9 & 89.7 & 91.7 & 95.4 \\
pcb2        & 84.3 & 67.5 & 85.5 & 78.6 & 87.8 & 76.6 & 92.6 \\
pcb3        & 87.0 & 83.3 & 86.9 & 80.3 & 85.8 & 82.1 & 83.9 \\
pcb4        & 95.6 & 87.6 & 91.2 & 97.8 & 97.4 & 95.3 & 99.1 \\
pipe fryum  & 96.4 & 98.5 & 99.0 & 98.6 & 99.5 & 97.9 & 99.1 \\
\midrule
mean        & 85.3 & 87.3 & 90.6 & 89.1 & 93.3 & \underline{93.6} & \textbf{94.0} \\
\bottomrule
\end{tabular}
\end{table*}
\begin{table*}[t]
\centering
\caption{Subset-wise performance comparison of the 1-shot setting for pAUROC on VisA.}
\label{tab:visa_1shot_pauroc}
\setlength{\tabcolsep}{6pt}
\renewcommand{\arraystretch}{0.85}
\begin{tabular}{lccccc cc}
\toprule
VisA & \multicolumn{7}{c}{1-shot} \\
\cmidrule(lr){2-8}
pAUROC & PatchCore & WinCLIP & AnomalyGPT & PromptAD & KAG-prompt & FocusPatch-AD & DriftAD (\textbf{Ours}) \\
\midrule
candle      & 97.2 & 97.4 & 98.2 & 95.8 & 98.6 & 98.1 & 98.8 \\
capsules    & 93.2 & 96.4 & 97.6 & 95.4 & 97.5 & 97.0 & 98.8 \\
cashew      & 98.1 & 98.5 & 96.5 & 99.1 & 96.8 & 98.0 & 96.6 \\
chewinggum  & 96.9 & 98.6 & 99.1 & 99.1 & 99.2 & 98.6 & 99.2 \\
fryum       & 93.3 & 96.4 & 92.7 & 95.4 & 93.8 & 95.4 & 95.0 \\
macaroni1   & 95.2 & 96.4 & 97.6 & 97.8 & 98.6 & 97.6 & 98.7 \\
macaroni2   & 89.1 & 96.8 & 95.3 & 96.6 & 97.3 & 96.6 & 97.7 \\
pcb1        & 96.1 & 96.6 & 97.9 & 96.6 & 97.4 & 97.4 & 98.8 \\
pcb2        & 95.4 & 93.0 & 92.5 & 93.5 & 94.0 & 93.9 & 95.5 \\
pcb3        & 96.2 & 94.3 & 94.6 & 95.9 & 95.6 & 95.7 & 95.7 \\
pcb4        & 95.6 & 94.0 & 94.9 & 95.5 & 96.9 & 94.7 & 95.5 \\
pipe fryum  & 98.8 & 98.3 & 98.0 & 99.1 & 98.4 & 98.9 & 98.6 \\
\midrule
mean        & 95.4 & 96.4 & 96.2 & 96.7 & \underline{97.0} & 96.8 & \textbf{97.4} \\
\bottomrule
\end{tabular}
\end{table*}
\begin{table*}[t]
\centering
\caption{Subset-wise performance comparison of the 2-shot setting for pAUROC on VisA.}
\label{tab:visa_2shot_pauroc}
\setlength{\tabcolsep}{6pt}
\renewcommand{\arraystretch}{0.85}
\begin{tabular}{lccccc cc}
\toprule
VisA & \multicolumn{7}{c}{2-shot} \\
\cmidrule(lr){2-8}
pAUROC & PatchCore & WinCLIP & AnomalyGPT & PromptAD & KAG-prompt & FocusPatch-AD & DriftAD {\textbf{Ours}} \\
\midrule
candle      & 97.7 & 97.7 & 98.3 & 95.9 & 98.8 & 97.1 & 99.0 \\
capsules    & 94.0 & 96.8 & 98.3 & 96.1 & 98.4 & 94.4 & 99.1 \\
cashew      & 98.2 & 98.5 & 95.7 & 99.2 & 98.2 & 98.5 & 98.2 \\
chewinggum  & 96.6 & 98.6 & 98.7 & 99.2 & 99.2 & 98.9 & 99.2 \\
fryum       & 94.0 & 97.0 & 93.6 & 96.4 & 93.8 & 96.8 & 95.1 \\
macaroni1   & 96.0 & 96.5 & 96.8 & 98.3 & 98.9 & 98.4 & 99.0 \\
macaroni2   & 90.2 & 96.8 & 96.3 & 97.2 & 97.4 & 96.8 & 97.8 \\
pcb1        & 97.6 & 97.0 & 96.5 & 96.9 & 98.0 & 96.7 & 98.9 \\
pcb2        & 96.0 & 93.9 & 93.0 & 94.8 & 95.2 &  94.3 & 96.7 \\
pcb3        & 97.1 & 95.1 & 95.9 & 96.1 & 95.6 & 95.7 & 95.6 \\
pcb4        & 96.2 & 95.6 & 95.3 & 95.6 & 97.3 & 96.2 &  96.4 \\
pipe fryum  & 99.1 & 98.5 & 98.2 & 99.4 & 98.4 & 98.7 & 98.6 \\
\midrule
mean        & 96.1 & 96.8 & 96.4 & 97.1 & 97.4 & \underline{97.7} & \textbf{97.8} \\
\bottomrule
\end{tabular}
\end{table*}
\begin{table*}[t]
\centering
\caption{Subset-wise performance comparison of the 4-shot setting for pAUROC on VisA.}
\label{tab:visa_4shot_pauroc}
\setlength{\tabcolsep}{6pt}
\renewcommand{\arraystretch}{0.85}
\begin{tabular}{lccccc cc}
\toprule
VisA & \multicolumn{7}{c}{4-shot} \\
\cmidrule(lr){2-8}
pAUROC & PatchCore & WinCLIP & AnomalyGPT & PromptAD & KAG-prompt & FocusPatch-AD & DriftAD (\textbf{Ours}) \\
\midrule
candle      & 97.9 & 97.8 & 98.6 & 96.0 & 98.8 & 95.4 & 99.0 \\
capsules    & 94.8 & 97.1 & 98.2 & 96.8 & 98.5 & 97.5 & 99.2 \\
cashew      & 98.3 & 98.7 & 96.9 & 99.2 & 98.3 & 98.2 & 98.2 \\
chewinggum  & 96.8 & 98.5 & 98.4 & 99.2 & 99.2 & 97.8 & 99.1 \\
fryum       & 94.2 & 97.1 & 94.3 & 96.6 & 94.7 & 95.8 & 95.6 \\
macaroni1   & 97.0 & 97.0 & 97.7 & 98.2 & 99.1 & 98.2 & 99.2 \\
macaroni2   & 93.9 & 97.3 & 96.6 & 97.0 & 97.4 & 97.0 & 97.6 \\
pcb1        & 98.1 & 98.1 & 96.8 & 98.2 & 97.9 & 98.4 & 98.9 \\
pcb2        & 96.6 & 94.6 & 93.2 & 95.3 & 96.1 & 95.9 & 97.2 \\
pcb3        & 97.4 & 95.8 & 96.1 & 96.8 & 96.4 & 96.3 & 96.5 \\
pcb4        & 97.0 & 96.1 & 95.6 & 96.2 & 97.6 & 96.7 & 96.7 \\
pipe fryum  & 99.1 & 98.7 & 98.5 & 99.3 & 98.5 & 99.0 & 98.7 \\
\midrule
mean        & 96.8 & 97.2 & 96.7 & 97.4 & 97.7 & \underline{97.9} & \textbf{98.0} \\
\bottomrule
\end{tabular}
\end{table*}
\begin{table*}[t]
\centering
\caption{Subset-wise performance comparison of the 1-shot setting for AUPR on VisA.}
\label{tab:visa_1shot_aupr}
\setlength{\tabcolsep}{5pt}
\renewcommand{\arraystretch}{0.85}
\begin{tabular}{lcccccc c}
\toprule
VisA & \multicolumn{7}{c}{1-shot} \\
\cmidrule(lr){2-8}
AUPR & PatchCore & WinCLIP & AnomalyGPT & PromptAD & KAG-prompt & FocusPatch-AD & DriftAD \textbf{(Ours)} \\
\midrule
candle      & 86.6 & 93.6 & 86.4 & 93.7 & 96.8 & 93.6 & 94.8 \\
capsules    & 72.3 & 89.9 & 91.4 & 90.1 & 94.1 & 89.7 & 94.7 \\
cashew      & 94.6 & 97.2 & 96.3 & 97.6 & 97.6 & 96.4 & 98.5 \\
chewinggum  & 98.9 & 99.0 & 99.1 & 99.1 & 99.4 & 98.4 & 99.8 \\
fryum       & 87.6 & 94.7 & 96.4 & 93.8 & 97.5 & 90.4 & 97.0 \\
macaroni1   & 67.8 & 84.9 & 92.8 & 86.3 & 97.0 & 89.5 & 96.9 \\
macaroni2   & 54.9 & 68.4 & 87.1 & 72.5 & 88.1 & 71.6 & 88.9 \\
pcb1        & 72.1 & 76.5 & 76.0 & 88.0 & 81.0 & 85.7 & 94.5 \\
pcb2        & 84.4 & 64.9 & 76.9 & 75.4 & 86.8 & 71.0 & 90.5 \\
pcb3        & 84.6 & 73.5 & 79.9 & 75.2 & 83.2 & 82.7 & 84.5 \\
pcb4        & 92.8 & 78.5 & 81.9 & 90.5 & 97.5 & 90.9 & 98.5 \\
pipe fryum  & 95.4 & 98.6 & 99.7 & 98.3 & 99.8 & 98.2 & 99.6 \\
\midrule
mean        & 82.8 & 85.1 & 88.7 & 88.4 & \underline{93.2} & 88.2 & \textbf{94.8} \\
\bottomrule
\end{tabular}
\end{table*}
\begin{table*}[t]
\centering
\caption{Subset-wise performance comparison of the 2-shot setting for AUPR on VisA.}
\label{tab:visa_2shot_aupr}
\setlength{\tabcolsep}{5pt}
\renewcommand{\arraystretch}{0.85}
\begin{tabular}{lcccccc c}
\toprule
VisA & \multicolumn{7}{c}{2-shot} \\
\cmidrule(lr){2-8}
AUPR & PatchCore & WinCLIP & AnomalyGPT & PromptAD & KAG-prompt & FocusPatch-AD & DriftAD \textbf{(Ours)} \\
\midrule
candle      & 86.8 & 95.1 & 83.2 & 93.6 & 95.3 & 95.3 & 94.3 \\
capsules    & 73.6 & 88.9 & 94.3 & 88.3 & 95.0 & 89.7 & 96.5 \\
cashew      & 96.9 & 97.3 & 96.1 & 97.4 & 97.3 & 95.3 & 97.7 \\
chewinggum  & 99.1 & 98.9 & 99.1 & 98.4 & 99.5 & 98.3 & 99.6 \\
fryum       & 92.1 & 95.8 & 97.5 & 96.0 & 97.9 & 93.6 & 98.0 \\
macaroni1   & 74.9 & 84.7 & 92.8 & 91.1 & 97.0 & 86.7 & 96.1 \\
macaroni2   & 57.2 & 70.4 & 88.6 & 84.7 & 89.8 & 76.6 & 88.2 \\
pcb1        & 72.6 & 78.3 & 78.7 & 80.9 & 86.7 & 89.6 & 91.9 \\
pcb2        & 86.6 & 65.8 & 75.8 & 73.0 & 85.2 & 74.2 & 92.1 \\
pcb3        & 86.1 & 80.9 & 80.4 & 82.8 & 90.0 & 82.5 & 86.9 \\
pcb4        & 93.2 & 72.5 & 81.8 & 94.5 & 96.8 & 96.6 & 98.9 \\
pipe fryum  & 96.8 & 99.0 & 99.7 & 99.1 & 99.9 & 99.6 & 99.6 \\
\midrule
mean        & 84.8 & 85.8 & 89.0 & 90.0 & \underline{94.2} & 89.8 & \textbf{95.0} \\
\bottomrule
\end{tabular}
\end{table*}
\begin{table*}[t]
\centering
\caption{Subset-wise performance comparison of the 4-shot setting for AUPR on VisA.}
\label{tab:visa_4shot_aupr}
\setlength{\tabcolsep}{5pt}
\renewcommand{\arraystretch}{0.85}
\begin{tabular}{lcccccc c}
\toprule
VisA & \multicolumn{7}{c}{4-shot} \\
\cmidrule(lr){2-8}
AUPR & PatchCore & WinCLIP & AnomalyGPT & PromptAD & KAG-prompt & FocusPatch-AD & DriftAD \textbf{(Ours)} \\
\midrule
candle      & 88.9 & 95.3 & 86.4 & 92.9 & 95.7 & 91.2 & 95.0 \\
capsules    & 78.4 & 91.5 & 94.3 & 89.8 & 95.0 & 92.2 & 95.7 \\
cashew      & 96.5 & 97.7 & 97.3 & 97.0 & 98.6 & 97.9 & 98.1 \\
chewinggum  & 99.3 & 99.0 & 99.2 & 98.5 & 99.3 & 98.8 & 99.7 \\
fryum       & 95.0 & 96.0 & 97.9 & 93.6 & 97.8 & 94.8 & 97.9 \\
macaroni1   & 82.1 & 86.5 & 92.7 & 98.2 & 98.0 & 86.9 & 97.9 \\
macaroni2   & 60.2 & 69.6 & 89.0 & 82.2 & 89.2 & 80.5 & 85.4 \\
pcb1        & 81.0 & 87.7 & 77.1 & 90.1 & 87.6 & 91.5 & 94.8 \\
pcb2        & 86.2 & 71.3 & 86.7 & 75.3 & 90.7 & 76.7 & 94.1 \\
pcb3        & 88.3 & 84.8 & 88.3 & 83.5 & 86.9 & 85.7 & 86.1 \\
pcb4        & 94.9 & 85.6 & 87.1 & 97.5 & 97.0 & 94.2 & 99.1 \\
pipe fryum  & 98.3 & 99.2 & 99.5 & 99.3 & 99.8 & 99.1 & 99.7 \\
\midrule
mean        & 87.5 & 88.8 & 91.3 & 91.5 & \underline{94.6} & 90.8 & \textbf{95.3} \\
\bottomrule
\end{tabular}
\end{table*}
\begin{table*}[t]
\centering
\caption{Subset-wise performance comparison of the 1-shot setting for PRO on VisA.}
\label{tab:visa_1shot_pro}
\setlength{\tabcolsep}{5pt}
\renewcommand{\arraystretch}{0.85}
\begin{tabular}{lcccccc c}
\toprule
VisA & \multicolumn{7}{c}{1-shot} \\
\cmidrule(lr){2-8}
PRO & PatchCore & WinCLIP & AnomalyGPT & PromptAD & KAG-prompt & FocusPatch-AD & DriftAD \textbf{(Ours)} \\
\midrule
candle      & 92.6 & 94.0 & 91.7 & 91.8 & 92.5 & 90.3 & 93.5 \\
capsules    & 66.6 & 73.6 & 82.4 & 70.0 & 78.9 & 61.4 & 83.2 \\
cashew      & 90.8 & 91.1 & 90.5 & 92.3 & 92.7 & 85.7 & 92.9 \\
chewinggum  & 78.2 & 91.0 & 87.3 & 89.8 & 88.2 & 77.9 & 88.6 \\
fryum       & 78.7 & 89.1 & 83.1 & 83.5 & 81.6 & 76.5 & 84.2 \\
macaroni1   & 83.4 & 84.6 & 87.5 & 87.5 & 92.0 & 84.2 & 92.5 \\
macaroni2   & 66.0 & 89.3 & 82.6 & 80.6 & 87.7 & 79.4 & 90.0 \\
pcb1        & 79.0 & 82.5 & 83.1 & 89.2 & 85.8 & 89.3 & 87.0 \\
pcb2        & 80.9 & 73.6 & 64.9 & 79.3 & 67.7 & 71.6 & 73.0 \\
pcb3        & 78.1 & 79.5 & 69.9 & 84.0 & 75.7 & 79.3 & 80.1 \\
pcb4        & 77.9 & 76.6 & 75.9 & 78.8 & 83.8 & 74.0 & 79.9 \\
pipe fryum  & 93.6 & 96.1 & 95.9 & 95.2 & 95.6 & 94.1 & 95.9 \\
\midrule
mean        & 80.5 & 85.1 & 82.9 & 85.1 & \underline{85.2} & 80.3 & \textbf{86.7} \\
\bottomrule
\end{tabular}
\end{table*}
\begin{table*}[t]
\centering
\caption{Subset-wise performance comparison of the 2-shot setting for PRO on VisA.}
\label{tab:visa_2shot_pro}
\setlength{\tabcolsep}{5pt}
\renewcommand{\arraystretch}{0.85}
\begin{tabular}{lcccccc c}
\toprule
VisA & \multicolumn{7}{c}{2-shot} \\
\cmidrule(lr){2-8}
PRO & PatchCore & WinCLIP & AnomalyGPT & PromptAD & KAG-prompt & FocusPatch-AD & DriftAD \textbf{(Ours)} \\
\midrule
candle      & 93.4 & 94.2 & 91.8 & 91.6 & 92.7 & 93.0 & 93.3 \\
capsules    & 67.9 & 75.9 & 85.7 & 70.8 & 85.9 & 82.2 & 88.0 \\
cashew      & 91.4 & 90.4 & 89.8 & 92.7 & 92.5 & 89.6 & 92.0 \\
chewinggum  & 78.0 & 90.9 & 85.4 & 87.8 & 88.0 & 84.0 & 88.1 \\
fryum       & 81.4 & 89.3 & 83.2 & 86.2 & 84.2 & 81.7 & 86.0 \\
macaroni1   & 86.2 & 85.2 & 87.4 & 90.6 & 92.5 & 91.3 & 93.4 \\
macaroni2   & 67.2 & 88.6 & 83.1 & 82.7 & 87.1 & 82.9 & 88.2 \\
pcb1        & 86.1 & 83.8 & 82.9 & 90.2 & 87.0 & 87.3 & 85.5 \\
pcb2        & 82.9 & 76.2 & 64.8 & 79.3 & 73.9 & 78.9 & 81.3 \\
pcb3        & 82.2 & 82.3 & 72.4 & 84.7 & 75.5 & 81.5 & 78.6 \\
pcb4        & 79.5 & 81.7 & 77.8 & 78.3 & 85.8 & 85.1 & 83.5 \\
pipe fryum  & 94.5 & 96.2 & 95.9 & 94.8 & 95.8 & 95.3 & 95.6 \\
\midrule
mean        & 82.6 & 86.2 & 83.4 & 85.8 & \underline{86.7} & 86.1 & \textbf{87.8} \\
\bottomrule
\end{tabular}
\end{table*}
\begin{table*}[t]
\centering
\caption{Subset-wise performance comparison of the 4-shot setting for PRO on VisA.}
\label{tab:visa_4shot_pro}
\setlength{\tabcolsep}{5pt}
\renewcommand{\arraystretch}{0.85}
\begin{tabular}{lcccccc c}
\toprule
VisA & \multicolumn{7}{c}{4-shot} \\
\cmidrule(lr){2-8}
PRO & PatchCore & WinCLIP & AnomalyGPT & PromptAD & KAG-prompt & FocusPatch-AD & DriftAD \textbf{(Ours)} \\
\midrule
candle      & 94.1 & 94.4 & 93.1 & 90.6 & 93.4 & 91.4 & 94.3 \\
capsules    & 69.0 & 77.0 & 85.1 & 72.4 & 85.8 & 77.2 & 88.2 \\
cashew      & 92.1 & 91.3 & 90.8 & 92.8 & 92.1 & 89.6 & 92.0 \\
chewinggum  & 79.3 & 91.0 & 85.1 & 89.4 & 87.9 & 86.3 & 87.4 \\
fryum       & 81.0 & 89.7 & 83.1 & 80.3 & 84.6 & 84.8 & 85.0 \\
macaroni1   & 89.6 & 86.8 & 90.7 & 91.5 & 93.7 & 92.6 & 94.6 \\
macaroni2   & 78.3 & 90.5 & 84.9 & 87.2 & 87.6 & 90.9 & 87.2 \\
pcb1        & 88.1 & 87.9 & 83.1 & 90.2 & 87.2 & 90.8 & 84.4 \\
pcb2        & 83.7 & 78.0 & 71.3 & 76.3 & 77.7 & 74.1 & 83.9 \\
pcb3        & 84.4 & 84.2 & 73.1 & 85.0 & 78.6 & 84.3 & 82.2 \\
pcb4        & 83.5 & 84.2 & 79.1 & 83.4 & 87.0 & 82.8 & 84.7 \\
pipe fryum  & 95.0 & 96.6 & 95.4 & 95.3 & 95.9 & 94.4 & 95.9 \\
\midrule
mean        & 84.9 & 87.6 & 84.6 & 86.2 & \underline{87.6} & 86.6 & \textbf{88.3} \\
\bottomrule
\end{tabular}
\end{table*}

\end{document}